# ParamILS: An Automatic Algorithm Configuration Framework

**Frank Hutter**                                    HUTTER@CS.UBC.CA
**Holger H. Hoos**                                   HOOS@CS.UBC.CA
**Kevin Leyton-Brown**                               KEVINLB@CS.UBC.CA
*University of British Columbia, 2366 Main Mall*
*Vancouver, BC, V6T1Z4, Canada*

**Thomas Stützle**                                   STUETZLE@ULB.AC.BE
*Université Libre de Bruxelles, CoDE, IRIDIA*
*Av. F. Roosevelt 50 B-1050 Brussels, Belgium*

## Abstract

The identification of performance-optimizing parameter settings is an important part of the development and application of algorithms. We describe an automatic framework for this algorithm configuration problem. More formally, we provide methods for optimizing a target algorithm's performance on a given class of problem instances by varying a set of ordinal and/or categorical parameters. We review a family of local-search-based algorithm configuration procedures and present novel techniques for accelerating them by adaptively limiting the time spent for evaluating individual configurations. We describe the results of a comprehensive experimental evaluation of our methods, based on the configuration of prominent complete and incomplete algorithms for SAT. We also present what is, to our knowledge, the first published work on automatically configuring the CPLEX mixed integer programming solver. All the algorithms we considered had default parameter settings that were manually identified with considerable effort. Nevertheless, using our automated algorithm configuration procedures, we achieved substantial and consistent performance improvements.

## 1. Introduction

Many high-performance algorithms have parameters whose settings control important aspects of their behaviour. This is particularly the case for heuristic procedures used for solving computationally hard problems.[1] As an example, consider CPLEX, a commercial solver for mixed integer programming problems.[2] CPLEX version 10 has about 80 parameters that affect the solver's search mechanism and can be configured by the user to improve performance. There are many acknowledgements in the literature that finding performance-optimizing parameter configurations of heuristic algorithms often requires considerable effort (see, e.g., Gratch & Chien, 1996; Johnson, 2002; Diao, Eskesen, Froehlich, Hellerstein, Spainhower & Surendra, 2003; Birattari, 2004; Adenso-Diaz & Laguna, 2006). In many cases, this tedious task is performed manually in an ad-hoc way. Automating this task is of high practical relevance in several contexts.

- **Development of complex algorithms** Setting the parameters of a heuristic algorithm is a highly labour-intensive task, and indeed can consume a large fraction of overall development time. The use of automated algorithm configuration methods can lead to significant time savings and potentially achieve better results than manual, ad-hoc methods.

---

1. Our use of the term 'heuristic algorithm' includes methods without provable performance guarantees as well as methods that have such guarantees, but nevertheless make use of heuristic mechanisms. In the latter case, the use of heuristic mechanisms often results in empirical performance far better than the bounds guaranteed by rigorous theoretical analysis.
2. http://www.ilog.com/products/cplex/

- **Empirical studies, evaluations, and comparisons of algorithms** A central question in comparing heuristic algorithms is whether one algorithm outperforms another because it is fundamentally superior, or because its developers more successfully optimized its parameters (Johnson, 2002). Automatic algorithm configuration methods can mitigate this problem of unfair comparisons and thus facilitate more meaningful comparative studies.

- **Practical use of algorithms** The ability of complex heuristic algorithms to solve large and hard problem instances often depends critically on the use of suitable parameter settings. End users often have little or no knowledge about the impact of an algorithm's parameter settings on its performance, and thus simply use default settings. Even if it has been carefully optimized on a standard benchmark set, such a default configuration may not perform well on the particular problem instances encountered by a user. Automatic algorithm configuration methods can be used to improve performance in a principled and convenient way.

A wide variety of strategies for automatic algorithm configuration have been explored in the literature. Briefly, these include exhaustive enumeration, hill-climbing (Gratch & Dejong, 1992), beam search (Minton, 1993), genetic algorithms (Terashima-Marín, Ross & Valenzuela-Réndon, 1999), experimental design approaches (Coy, Golden, Runger & Wasil, 2001), sequential parameter optimization (Bartz-Beielstein, 2006), racing algorithms (Birattari, Stützle, Paquete & Varrentrapp, 2002; Birattari, 2004; Balaprakash, Birattari & Stützle, 2007), and combinations of fractional experimental design and local search (Adenso-Diaz & Laguna, 2006). We discuss this and other related work more extensively in Section 9. Here, we note that while some other authors refer to the optimization of an algorithm's performance by setting its (typically few and numerical) parameters as *parameter tuning*, we favour the term *algorithm configuration* (or simply, *configuration*). This is motivated by the fact that we are interested in methods that can deal with a potentially large number of parameters, each of which can be numerical, ordinal (e.g., low, medium, or high) or categorical (e.g., choice of heuristic). Categorical parameters can be used to select and combine discrete building blocks of an algorithm (e.g., preprocessing and variable ordering heuristics); consequently, our general view of algorithm configuration includes the automated construction of a heuristic algorithm from such building blocks. To the best of our knowledge, the methods discussed in this article are yet the only general ones available for the configuration of algorithms with many categorical parameters.

We now give an overview of what follows and highlight our main contributions. After formally stating the algorithm configuration problem in Section 2, in Section 3 we describe ParamILS (first introduced by Hutter, Hoos & Stützle, 2007), a versatile stochastic local search approach for automated algorithm configuration, and two of its instantiations, BasicILS and FocusedILS.

We then introduce *adaptive capping* of algorithm runs, a novel technique that can be used to enhance search-based algorithm configuration procedures independently of the underlying search strategy (Section 4). Adaptive capping is based on the idea of avoiding unnecessary runs of the algorithm to be configured by developing bounds on the performance measure to be optimized. We present a trajectory-preserving variant and a heuristic extension of this technique. After discussing experimental preliminaries in Section 5, in Section 6 we present empirical evidence showing that adaptive capping speeds up both BasicILS and FocusedILS. We also show that BasicILS outperforms random search and a simple local search, as well as further evidence that FocusedILS outperforms BasicILS.

We present extensive evidence that ParamILS can find substantially improved parameter configurations of complex and highly optimized algorithms. In particular, we apply our automatic algorithm configuration procedures to the aforementioned commercial optimization tool CPLEX, one of the most powerful, widely used and complex optimization algorithms we are aware of. As stated in the CPLEX user manual (version 10.0, page 247), "A great deal of algorithmic development effort has been devoted to establishing default ILOG CPLEX parameter settings that achieve good performance on a wide variety of MIP models." We demonstrate consistent improvements over this default parameter configuration for a wide range of practically relevant instance distributions. In some cases, we were able to achieve an average speedup of over an order of magnitude on previously-unseen test instances (Section 7). We believe that these are the first results to be published on automatically configuring CPLEX or any other piece of software of comparable complexity.

In Section 8 we review a wide range of (separately-published) ParamILS applications. Specifically, we survey work that has considered the optimization of complete and incomplete heuristic search algorithms for the problems of propositional satisfiability (SAT), most probable explanation (MPE), protein folding, university time-tabling, and algorithm configuration itself. In three of these cases, ParamILS was an integral part of the algorithm design process and allowed the exploration of very large design spaces. This could not have been done effectively in a manual way or by any other existing automated method. Thus, automated algorithm configuration in general and ParamILS in particular enables a new way of (semi-)automatic design of algorithms from components.

Section 9 presents related work and, finally, Section 10 offers discussion and conclusions. Here we distill the common patterns that helped ParamILS to succeed in its various applications. We also give advice to practitioners who would like to apply automated algorithm configuration in general and ParamILS in particular, and identify promising avenues of research for future work.

## 2. Problem Statement and Notation

The algorithm configuration problem we consider in this work can be informally stated as follows: given an algorithm, a set of parameters for the algorithm and a set of input data, find parameter values under which the algorithm achieves the best possible performance on the input data.

To avoid potential confusion between algorithms whose performance is optimized and algorithms used for carrying out that optimization task, we refer to the former as *target algorithms* and to the latter as *configuration procedures* (or simply *configurators*). This setup is illustrated in Figure 1. Different algorithm configuration problems have also been considered in the literature, including setting parameters on a per-instance basis and adapting the parameters while the algorithm is running. We defer a discussion of these approaches to Section 9.

In the following, we define the algorithm configuration problem more formally and introduce notation that we will use throughout this article. Let $\mathcal{A}$ denote an algorithm, and let $p_1, \ldots, p_k$ be parameters of $\mathcal{A}$. Denote the domain of possible values for each parameter $p_i$ as $\Theta_i$. Throughout this work, we assume that all parameter domains are finite sets. This assumption can be met by discretizing all numerical parameters to a finite number of values. Furthermore, while parameters may be ordered, we do not exploit such ordering relations. Thus, we effectively assume that all parameters are finite and categorical.[3]

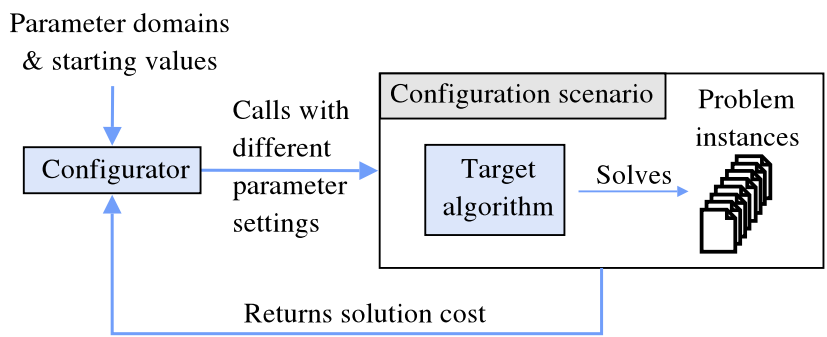

Figure 1: A configuration scenario includes an algorithm to be configured and a collection of problem instances. A configuration procedure executes the target algorithm with specified parameter settings on some or all of the instances, receives information about the performance of these runs, and uses this information to decide which subsequent parameter configurations to evaluate.

Our problem formulation allows us to express conditional parameter dependencies (for example, one algorithm parameter might be used to select among search heuristics, with each heuristic's behaviour controlled by further parameters). In this case, the values of these further parameters are irrelevant if the heuristic is not selected. ParamILS exploits this and effectively searches the space of equivalence classes in parameter configuration space. In addition, our formulation supports constraints on feasible combinations of parameter values. We use $\Theta \subseteq \Theta_1 \times \ldots \times \Theta_k$ to denote the space of all feasible parameter configurations, and $\mathcal{A}(\boldsymbol{\theta})$ denoting the instantiation of algorithm $\mathcal{A}$ with parameter configuration $\boldsymbol{\theta} \in \Theta$.

Let $\mathcal{D}$ denote a probability distribution over a space $\Pi$ of problem instances, and denote an element of $\Pi$ as $\pi$. $\mathcal{D}$ may be given implicitly, as through a random instance generator or a distribution over such generators. It is also possible (and indeed common) for $\Pi$ to consist of a finite sample of instances; in this case, we define $\mathcal{D}$ as the uniform distribution over $\Pi$.

There are many ways of measuring an algorithm's performance. For example, we might be interested in minimizing computational resources consumed by the given algorithm (such as runtime, memory or communication bandwidth), or in maximizing the quality of the solution found. Since high-performance algorithms for computationally-challenging problems are often randomized, their behaviour can vary significantly between multiple runs. Thus, an algorithm will not always achieve the same performance, even when run repeatedly with fixed parameters on a single problem instance. Our overall goal must therefore be to choose parameter settings that minimize some cost statistic of the algorithm's performance across the input data. We denote this statistic as $c(\boldsymbol{\theta})$. For example, we might aim to minimize mean runtime or median solution cost.

With this intuition in mind, we now define the algorithm configuration problem formally.

**Definition 1** (Algorithm Configuration Problem). *An instance of the* algorithm configuration *problem is a 6-tuple* $\langle \mathcal{A}, \Theta, \mathcal{D}, \kappa_{max}, o, m \rangle$, *where:*

- $\mathcal{A}$ *is a parameterized algorithm;*
- $\Theta$ *is the parameter configuration space of* $\mathcal{A}$;

---

3. We are currently extending our algorithm configuration procedures to natively support other parameter types.

- $\mathcal{D}$ *is a distribution over problem instances with domain* $\Pi$;

- $\kappa_{max}$ *is a cutoff time (or* captime*), after which each run of* $\mathcal{A}$ *will be terminated if still running;*

- $o$ *is a function that measures the observed cost of running* $\mathcal{A}(\boldsymbol{\theta})$ *on an instance* $\pi \in \Pi$ *with captime* $\kappa \in \mathbb{R}$ *(examples are runtime for solving the instance, or cost of the solution found)*

- $m$ *is a statistical population parameter (such as expectation, median, or variance).*

Any parameter configuration $\boldsymbol{\theta} \in \boldsymbol{\Theta}$ is a candidate solution of the algorithm configuration problem. For each parameter configuration $\boldsymbol{\theta}$, $O_{\boldsymbol{\theta}}$ denotes the *distribution of costs* induced by function $o$, applied to instances $\pi$ drawn from distribution $\mathcal{D}$ and multiple independent runs for randomized algorithms, using captime $\kappa = \kappa_{max}$. The *cost* of a candidate solution $\boldsymbol{\theta}$ is defined as

$$c(\boldsymbol{\theta}) := m(O_{\boldsymbol{\theta}}), \tag{1}$$

the statistical population parameter $m$ of the cost distribution $O_{\boldsymbol{\theta}}$. An optimal solution, $\boldsymbol{\theta}^*$, minimizes $c(\boldsymbol{\theta})$:

$$\boldsymbol{\theta}^* \in \underset{\boldsymbol{\theta} \in \boldsymbol{\Theta}}{\arg\min}\, c(\boldsymbol{\theta}). \tag{2}$$

An *algorithm configuration procedure* is a procedure for solving the algorithm configuration problem. Unfortunately, at least for the algorithm configuration problems considered in this article, we cannot optimize $c$ in closed form since we do not have access to an algebraic representation of the function. We denote the sequence of runs executed by a configurator as $\mathbf{R} = ((\boldsymbol{\theta}_1, \pi_1, s_1, \kappa_1, o_1), \ldots, (\boldsymbol{\theta}_n, \pi_n, s_n, \kappa_n, o_n))$. The $i$th run is described by five values:

- $\boldsymbol{\theta}_i \in \boldsymbol{\Theta}$ denotes the parameter configuration being evaluated;

- $\pi_i \in \Pi$ denotes the instance on which the algorithm is run;

- $s_i$ denotes the random number seed used in the run (we keep track of seeds to be able to block on them, see Section 5.1.2);

- $\kappa_i$ denotes the run's captime; and

- $o_i$ denotes the observed cost of the run

Note that each of $\boldsymbol{\theta}$, $\pi$, $s$, $\kappa$, and $o$ can vary from one element of $\mathbf{R}$ to the next, regardless of whether or not other elements are held constant. We denote the $i$th run of $\mathbf{R}$ as $\mathbf{R}[i]$, and the subsequence of runs using parameter configuration $\boldsymbol{\theta}$ (i.e., those runs with $\boldsymbol{\theta}_i = \boldsymbol{\theta}$) as $\mathbf{R}_{\boldsymbol{\theta}}$. The configuration procedures considered in this article compute empirical estimates of $c(\boldsymbol{\theta})$ based solely on $\mathbf{R}_{\boldsymbol{\theta}}$, but in principle other methods could be used. We compute these *cost estimates* both online, during runtime of a configurator, as well as offline, for evaluation purposes.

**Definition 2** (Cost Estimate)**.** *Given an algorithm configuration problem* $\langle \mathcal{A}, \boldsymbol{\Theta}, \mathcal{D}, \kappa_{max}, o, m \rangle$, *we define a* cost estimate *of a cost* $c(\boldsymbol{\theta})$ *based on a sequence of runs* $\mathbf{R} = ((\boldsymbol{\theta}_1, \pi_1, s_1, \kappa_1, o_1), \ldots, (\boldsymbol{\theta}_n, \pi_n, s_n, \kappa_n, o_n))$ *as* $\hat{c}(\boldsymbol{\theta}, \mathbf{R}) := \hat{m}(\{o_i \mid \boldsymbol{\theta}_i = \boldsymbol{\theta}\})$, *where* $\hat{m}$ *is the sample statistic analogue to the statistical population parameter* $m$.

For example, when $c(\boldsymbol{\theta})$ is the expected runtime over a distribution of instances and random number seeds, $\hat{c}(\boldsymbol{\theta}, \mathbf{R})$ is the sample mean runtime of runs $\mathbf{R}_{\boldsymbol{\theta}}$.

All configuration procedures in this paper are anytime algorithms, meaning that at all times they keep track of the configuration currently believed to have the lowest cost; we refer to this configuration as the *incumbent configuration*, or in short the *incumbent*, $\boldsymbol{\theta}_{inc}$. We evaluate a configurator's performance at time $t$ by means of its incumbent's training and test performance, defined as follows.

**Definition 3** (Training performance). *When at some time $t$ a configurator has performed a sequence of runs $\mathbf{R} = ((\boldsymbol{\theta}_1, \pi_1, s_1, \kappa_1, o_1), \ldots, (\boldsymbol{\theta}_n, \pi_n, s_n, \kappa_n, o_n))$ to solve an algorithm configuration problem $\langle \mathcal{A}, \boldsymbol{\Theta}, \mathcal{D}, \kappa_{max}, o, m \rangle$, and has thereby found incumbent configuration $\boldsymbol{\theta}_{inc}$, then its* training performance *at time $t$ is defined as the cost estimate $\hat{c}(\boldsymbol{\theta}_{inc}, \mathbf{R})$.*

The set of instances $\{\pi_1, \ldots, \pi_n\}$ discussed above is called the *training set*. While the true cost of a parameter configuration cannot be computed exactly, it can be estimated using training performance. However, the training performance of a configurator is a biased estimator of its incumbent's true cost, because the same instances are used for selecting the incumbent as for evaluating it. In order to achieve unbiased estimates during offline evaluation, we set aside a fixed set of instances $\{\pi'_1, \ldots, \pi'_T\}$ (called the *test set*) and random number seeds $\{s'_1, \ldots, s'_T\}$, both unknown to the configurator, and use these for evaluation.

**Definition 4** (Test performance). *At some time $t$, let a configurator's incumbent for an algorithm configuration problem $\langle \mathcal{A}, \boldsymbol{\Theta}, \mathcal{D}, \kappa_{max}, o, m \rangle$ be $\boldsymbol{\theta}_{inc}$ (this is found by means of executing a sequence of runs on the training set). Furthermore, let $\mathbf{R}' = ((\boldsymbol{\theta}_{inc}, \pi'_1, s'_1, \kappa_{max}, o_1), \ldots, (\boldsymbol{\theta}_{inc}, \pi'_T, s'_T, \kappa_{max}, o_T))$ be a sequence of runs on the $T$ instances and random number seeds in the test set (which is performed offline for evaluation purposes), then the configurator's* test performance *at time $t$ is defined as the cost estimate $\hat{c}(\boldsymbol{\theta}_{inc}, \mathbf{R}')$.*

Throughout this article, we aim to minimize expected runtime. (See Section 5.1.1 for a discussion of that choice.) Thus, a configurator's training performance is the mean runtime of the runs it performed with the incumbent. Its test performance is the mean runtime of the incumbent on the test set. Note that, while the configurator is free to use any $\kappa_i \leq \kappa_{max}$, test performance is always computed using the maximal captime, $\kappa_{max}$.

It is not obvious how an automatic algorithm configurator should choose runs in order to best minimize $c(\boldsymbol{\theta})$ within a given time budget. In particular, we have to make the following choices:

1. Which parameter configurations $\boldsymbol{\Theta}' \subseteq \boldsymbol{\Theta}$ should be evaluated?

2. Which problem instances $\Pi_{\boldsymbol{\theta}'} \subseteq \Pi$ should be used for evaluating each $\boldsymbol{\theta}' \in \boldsymbol{\Theta}'$, and how many runs should be performed on each instance?

3. Which cutoff time $\kappa_i$ should be used for each run?

Hutter, Hoos and Leyton-Brown (2009) considered this design space in detail, focusing on the tradeoff between the (fixed) number of problem instances to be used for the evaluation of each parameter configuration and the (fixed) cutoff time used for each run, as well as the interaction of these choices with the number of configurations that can be considered. In contrast, here, we study *adaptive* approaches for selecting the number of problem instances (Section 3.3) and the cutoff time for the evaluation of a parameter configuration (Section 4); we also study *which* configurations should be selected (Sections 3.1 and 6.2).

## 3. ParamILS: Iterated Local Search in Parameter Configuration Space

In this section, we address the first and most important of the previously mentioned dimensions of automated algorithm configuration, the search strategy, by describing an iterated local search framework called ParamILS. To start with, we fix the other two dimensions, using an unvarying benchmark set of instances and fixed cutoff times for the evaluation of each parameter configuration. Thus, the stochastic optimization problem of algorithm configuration reduces to a simple optimization problem, namely to find the parameter configuration that yields the lowest mean run-time on the given benchmark set. Then, in Section 3.3, we address the second question of how many runs should be performed for each configuration.

### 3.1 The ParamILS framework

Consider the following manual parameter optimization process:

1. begin with some initial parameter configuration;

2. experiment with modifications to single parameter values, accepting new configurations whenever they result in improved performance;

3. repeat step 2 until no single-parameter change yields an improvement.

This widely used procedure corresponds to a manually-executed local search in parameter configuration space. Specifically, it corresponds to an iterative first improvement procedure with a search space consisting of all possible configurations, an objective function that quantifies the performance achieved by the target algorithm with a given configuration, and a neighbourhood relation based on the modification of one single parameter value at a time (i.e., a "one-exchange" neighbourhood).

Viewing this manual procedure as a local search algorithm is advantageous because it suggests the automation of the procedure as well as its improvement by drawing on ideas from the stochastic local search community. For example, note that the procedure stops as soon as it reaches a local optimum (a parameter configuration that cannot be improved by modifying a single parameter value). A more sophisticated approach is to employ iterated local search (ILS; Lourenço, Martin & Stützle, 2002) to search for performance-optimizing parameter configurations. ILS is a prominent stochastic local search method that builds a chain of local optima by iterating through a main loop consisting of (1) a solution perturbation to escape from local optima, (2) a subsidiary local search procedure and (3) an acceptance criterion to decide whether to keep or reject a newly obtained candidate solution.

ParamILS (given in pseudocode as Algorithm 1) is an ILS method that searches parameter configuration space. It uses a combination of default and random settings for initialization, employs iterative first improvement as a subsidiary local search procedure, uses a fixed number ($s$) of random moves for perturbation, and always accepts better or equally-good parameter configurations, but re-initializes the search at random with probability $p_{restart}$.[4] Furthermore, it is based on a one-exchange neighbourhood, that is, we always consider changing only one parameter at a time. ParamILS deals with conditional parameters by excluding all configurations from the neighbourhood of a configuration $\boldsymbol{\theta}$ that differ only in a conditional parameter that is not relevant in $\boldsymbol{\theta}$.

### 3.2 The BasicILS Algorithm

In order to turn ParamILS as specified in Algorithm Framework 1 into an executable configuration procedure, it is necessary to instantiate the function *better* that determines which of two parameter settings should be preferred. We will ultimately propose several different ways of doing this. Here, we describe the simplest approach, which we call *BasicILS*. Specifically, we use the term BasicILS($N$) to refer to a ParamILS algorithm in which the function *better*($\boldsymbol{\theta}_1, \boldsymbol{\theta}_2$) is implemented as shown in Procedure 2: simply comparing estimates $\hat{c}_N$ of the cost statistics $c(\boldsymbol{\theta}_1)$ and $c(\boldsymbol{\theta}_2)$ that are based on $N$ runs each.

---

4. Our original parameter choices $\langle r, s, p_{restart} \rangle = \langle 10, 3, 0.01 \rangle$ (from Hutter et al., 2007) were somewhat arbitrary, though we expected performance to be quite robust with respect to these settings. We revisit this issue in Section 8.4.

---

**Algorithm Framework 1: ParamILS($\theta_0, r, p_{restart}, s$)**

Outline of iterated local search in parameter configuration space; the specific variants of ParamILS we study, **BasicILS(N)** and **FocusedILS**, are derived from this framework by instantiating procedure *better* (which compares $\theta, \theta' \in \Theta$). *BasicILS(N)* uses *better$_N$* (see Procedure 2), while FocusedILS uses *better$_{Foc}$* (see Procedure 3). The neighbourhood *Nbh($\theta$)* of a configuration $\theta$ is the set of all configurations that differ from $\theta$ in one parameter, excluding configurations differing in a conditional parameter that is not relevant in $\theta$.

---

**Input** : Initial configuration $\theta_0 \in \Theta$, algorithm parameters $r$, $p_{restart}$, and $s$.

**Output** : Best parameter configuration $\theta$ found.

**1** **for** $i = 1, \dots, r$ **do**

**2**     $\theta \leftarrow$ random $\theta \in \Theta$;

**3**     **if** better($\theta, \theta_0$) **then** $\theta_0 \leftarrow \theta$;

**4** $\theta_{ils} \leftarrow$ *IterativeFirstImprovement* ($\theta_0$);

**5** **while not** *TerminationCriterion()* **do**

**6**     $\theta \leftarrow \theta_{ils}$;

    // ===== *Perturbation*

**7**     **for** $i = 1, \dots, s$ **do** $\theta \leftarrow$ random $\theta' \in Nbh(\theta)$;

    // ===== *Basic local search*

**8**     $\theta \leftarrow$ *IterativeFirstImprovement* ($\theta$);

    // ===== *AcceptanceCriterion*

**9**     **if** better($\theta, \theta_{ils}$) **then** $\theta_{ils} \leftarrow \theta$;

**10**     **with probability** $p_{restart}$ **do** $\theta_{ils} \leftarrow$ random $\theta \in \Theta$;

**11** **return** overall best $\theta_{inc}$ found;

**12** **Procedure** *IterativeFirstImprovement* ($\theta$)

**13** **repeat**

**14**     $\theta' \leftarrow \theta$;

**15**     **foreach** $\theta'' \in Nbh(\theta')$ *in randomized order* **do**

**16**       **if** better($\theta'', \theta'$) **then** $\theta \leftarrow \theta''$; **break**;

**17** **until** $\theta' = \theta$;

**18** **return** $\theta$;

---

BasicILS($N$) is a simple and intuitive approach since it evaluates every parameter configuration by running it on the same $N$ training benchmark instances using the same random number seeds. Like many other related approaches (see, e.g., Minton, 1996; Coy et al., 2001; Adenso-Diaz & Laguna, 2006), it deals with the stochastic part of the optimisation problem by using an estimate based on a fixed training set of $N$ instances. When benchmark instances are very heterogeneous or

---

**Procedure 2: better$_N$($\theta_1, \theta_2$)**

Procedure used in BasicILS($N$) and RandomSearch($N$) to compare two parameter configurations. Procedure objective($\theta, N$) returns the user-defined objective achieved by $\mathcal{A}(\theta)$ on the first $N$ instances and keeps track of the incumbent solution, $\theta_{inc}$; it is detailed in Procedure 4 on page 279.

---

**Input** : Parameter configuration $\theta_1$, parameter configuration $\theta_2$

**Output** : True if $\theta_1$ does better than or equal to $\theta_2$ on the first $N$ instances; false otherwise

**Side Effect** : Adds runs to the global caches of performed algorithm runs $\mathbf{R}_{\theta_1}$ and $\mathbf{R}_{\theta_2}$; potentially updates the incumbent $\theta_{inc}$

**1** $\hat{c}_N(\theta_2) \leftarrow objective(\theta_2, N)$

**2** $\hat{c}_N(\theta_1) \leftarrow objective(\theta_1, N)$

**3** **return** $\hat{c}_N(\theta_1) \leq \hat{c}_N(\theta_2)$

---

when the user can identify a rather small "representative" subset of instances, this approach can find good parameter configurations with low computational effort.

### 3.3 FocusedILS: Adaptively Selecting the Number of Training Instances

The question of how to choose the number of training instances, $N$, in BasicILS($N$) has no straight-forward answer: optimizing performance using too small a training set leads to good training performance, but poor generalization to previously unseen test benchmarks. On the other hand, we clearly cannot evaluate every parameter configuration on an enormous training set—if we did, search progress would be unreasonably slow.

FocusedILS is a variant of ParamILS that deals with this problem by adaptively varying the number of training samples considered from one parameter configuration to another. We denote the number of runs available to estimate the cost statistic $c(\boldsymbol{\theta})$ for a parameter configuration $\boldsymbol{\theta}$ by $N(\boldsymbol{\theta})$. Having performed different numbers of runs using different parameter configurations, we face the question of comparing two parameter configurations $\boldsymbol{\theta}$ and $\boldsymbol{\theta}'$ for which $N(\boldsymbol{\theta}) \leq N(\boldsymbol{\theta}')$. One option would be simply to compute the empirical cost statistic based on the available number of runs for each configuration. However, this can lead to systematic biases if, for example, the first instances are easier than the average instance. Instead, we compare $\boldsymbol{\theta}$ and $\boldsymbol{\theta}'$ based on $N(\boldsymbol{\theta})$ runs on the same instances and seeds. This amounts to a blocking strategy, which is a straight-forward adaptation of a known variance reduction technique; see 5.1 for a more detailed discussion.

This approach to comparison leads us to a concept of domination. We say that $\boldsymbol{\theta}$ dominates $\boldsymbol{\theta}'$ when at least as many runs have been conducted on $\boldsymbol{\theta}$ as on $\boldsymbol{\theta}'$, and the performance of $\mathcal{A}(\boldsymbol{\theta})$ on the first $N(\boldsymbol{\theta}')$ runs is at least as good as that of $\mathcal{A}(\boldsymbol{\theta}')$ on all of its runs.

**Definition 5** (Domination). $\boldsymbol{\theta}_1$ *dominates* $\boldsymbol{\theta}_2$ *if and only if* $N(\boldsymbol{\theta}_1) \geq N(\boldsymbol{\theta}_2)$ *and* $\hat{c}_{N(\boldsymbol{\theta}_2)}(\boldsymbol{\theta}_1) \leq \hat{c}_{N(\boldsymbol{\theta}_2)}(\boldsymbol{\theta}_2)$.

Now we are ready to discuss the comparison strategy encoded in procedure $better_{Foc}(\boldsymbol{\theta}_1, \boldsymbol{\theta}_2)$, which is used by the FocusedILS algorithm (see Procedure 3). This procedure first acquires one additional sample for the configuration $i$ having smaller $N(\boldsymbol{\theta}_i)$, or one run for both configurations if they have the same number of runs. Then, it continues performing runs in this way until one configuration dominates the other. At this point it returns true if $\boldsymbol{\theta}_1$ dominates $\boldsymbol{\theta}_2$, and false otherwise. We also keep track of the total number of configurations evaluated since the last improving step (i.e., since the last time $better_{Foc}$ returned true); we denote this number as $B$. Whenever $better_{Foc}(\boldsymbol{\theta}_1, \boldsymbol{\theta}_2)$ returns true, we perform $B$ "bonus" runs for $\boldsymbol{\theta}_1$ and reset $B$ to 0. This mechanism ensures that we perform many runs with good configurations, and that the error made in every comparison of two configurations $\boldsymbol{\theta}_1$ and $\boldsymbol{\theta}_2$ decreases on expectation.

It is not difficult to show that in the limit, FocusedILS will sample every parameter configuration an unbounded number of times. The proof relies on the fact that, as an instantiation of ParamILS, FocusedILS performs random restarts with positive probability.

**Lemma 6** (Unbounded number of evaluations). *Let* $N(J, \boldsymbol{\theta})$ *denote the number of runs FocusedILS has performed with parameter configuration* $\boldsymbol{\theta}$ *at the end of ILS iteration* $J$ *to estimate* $c(\boldsymbol{\theta})$. *Then, for any constant* $K$ *and configuration* $\boldsymbol{\theta} \in \Theta$ *(with finite* $|\Theta|$*),* $\lim_{J \to \infty} P\left[N(J, \boldsymbol{\theta}) \geq K\right] = 1$.

*Proof.* After each ILS iteration of ParamILS, with probability $p_{restart} > 0$ a new configuration is picked uniformly at random, and with probability $1/|\Theta|$, this is configuration $\boldsymbol{\theta}$. The probability of

---

**Procedure 3:** $better_{Foc}(\boldsymbol{\theta}_1, \boldsymbol{\theta}_2)$

Procedure used in FocusedILS to compare two parameter configurations. Procedure $objective(\boldsymbol{\theta}, N)$ returns the user-defined objective achieved by $\mathcal{A}(\boldsymbol{\theta})$ on the first $N$ instances, keeps track of the incumbent solution, and updates $\mathbf{R}_{\boldsymbol{\theta}}$ (a global cache of algorithm runs performed with parameter configuration $\boldsymbol{\theta}$); it is detailed in Procedure 4 on page 279. For each $\boldsymbol{\theta}$, $N(\boldsymbol{\theta}) = \text{length}(\mathbf{R}_{\boldsymbol{\theta}})$. $B$ is a global counter denoting the number of configurations evaluated since the last improvement step.

---

**Input** : Parameter configuration $\boldsymbol{\theta}_1$, parameter configuration $\boldsymbol{\theta}_2$
**Output** : True if $\boldsymbol{\theta}_1$ dominates $\boldsymbol{\theta}_2$, false otherwise
**Side Effect**: Adds runs to the global caches of performed algorithm runs $\mathbf{R}_{\boldsymbol{\theta}_1}$ and $\mathbf{R}_{\boldsymbol{\theta}_2}$; updates the global counter $B$ of bonus runs, and potentially the incumbent $\boldsymbol{\theta}_{inc}$

**1** $B \leftarrow B + 1$
**2 if** $N(\boldsymbol{\theta}_1) \leq N(\boldsymbol{\theta}_2)$ **then**
**3** $\quad \boldsymbol{\theta}_{min} \leftarrow \boldsymbol{\theta}_1; \boldsymbol{\theta}_{max} \leftarrow \boldsymbol{\theta}_2$
**4** $\quad$ **if** $N(\boldsymbol{\theta}_1) = N(\boldsymbol{\theta}_2)$ **then** $B \leftarrow B + 1$
**5 else** $\boldsymbol{\theta}_{min} \leftarrow \boldsymbol{\theta}_2; \boldsymbol{\theta}_{max} \leftarrow \boldsymbol{\theta}_1$
**6 repeat**
**7** $\quad i \leftarrow N(\boldsymbol{\theta}_{min}) + 1$
**8** $\quad \hat{c}_i(\boldsymbol{\theta}_{max}) \leftarrow objective(\boldsymbol{\theta}_{max}, i)$ $\;$ // If $N(\boldsymbol{\theta}_{min}) = N(\boldsymbol{\theta}_{max})$, adds a new run to $\mathbf{R}_{\boldsymbol{\theta}_{max}}$.
**9** $\quad \hat{c}_i(\boldsymbol{\theta}_{min}) \leftarrow objective(\boldsymbol{\theta}_{min}, i)$ $\;$ // Adds a new run to $\mathbf{R}_{\boldsymbol{\theta}_{min}}$.
**10 until** $dominates(\boldsymbol{\theta}_1, \boldsymbol{\theta}_2)$ or $dominates(\boldsymbol{\theta}_2, \boldsymbol{\theta}_1)$
**11 if** $dominates(\boldsymbol{\theta}_1, \boldsymbol{\theta}_2)$ **then**
$\quad$ // ===== Perform B bonus runs.
**12** $\quad \hat{c}_{N(\boldsymbol{\theta}_1)+B}(\boldsymbol{\theta}_1) \leftarrow objective(\boldsymbol{\theta}_1, N(\boldsymbol{\theta}_1) + B)$ $\;$ // Adds B new runs to $\mathbf{R}_{\boldsymbol{\theta}_1}$.
**13** $\quad B \leftarrow 0$
**14** $\quad$ **return** true
**15 else return** false

**16 Procedure** dominates$(\boldsymbol{\theta}_1, \boldsymbol{\theta}_2)$
**17 if** $N(\boldsymbol{\theta}_1) < N(\boldsymbol{\theta}_2)$ **then return** false
**18 return** $objective(\boldsymbol{\theta}_1, N(\boldsymbol{\theta}_2)) \leq objective(\boldsymbol{\theta}_2, N(\boldsymbol{\theta}_2))$

---

visiting $\boldsymbol{\theta}$ in an ILS iteration is thus $p \geq \frac{p_{restart}}{|\boldsymbol{\Theta}|} > 0$. Hence, the number of runs performed with $\boldsymbol{\theta}$ is lower-bounded by a binomial random variable $\mathcal{B}(k; J, p)$. Then, for any constant $k < K$ we obtain $\lim_{J \to \infty} \mathcal{B}(k; J, p) = \lim_{J \to \infty} \binom{J}{k} p^k (1-p)^{J-k} = 0$. Thus, $\lim_{J \to \infty} P[N(J, \boldsymbol{\theta}) \geq K] = 1$. $\qquad \square$

**Definition 7** (Consistent estimator). *$\hat{c}_N(\boldsymbol{\theta})$ is a* consistent estimator *for $c(\boldsymbol{\theta})$ iff*

$$\forall \epsilon > 0 \; : \; \lim_{N \to \infty} P(|\hat{c}_N(\boldsymbol{\theta}) - c(\boldsymbol{\theta})| < \epsilon) = 1.$$

When $\hat{c}_N(\boldsymbol{\theta})$ is a consistent estimator of $c(\boldsymbol{\theta})$, cost estimates become more and more reliable as $N$ approaches infinity, eventually eliminating overconfidence and the possibility of mistakes in comparing two parameter configurations. This fact is captured in the following lemma.

**Lemma 8** (No mistakes for $N \to \infty$). *Let $\boldsymbol{\theta}_1, \boldsymbol{\theta}_2 \in \boldsymbol{\Theta}$ be any two parameter configurations with $c(\boldsymbol{\theta}_1) < c(\boldsymbol{\theta}_2)$. Then, for consistent estimators $\hat{c}_N$, $\lim_{N \to \infty} P(\hat{c}_N(\boldsymbol{\theta}_1) \geq \hat{c}_N(\boldsymbol{\theta}_2)) = 0$.*

*Proof.* Write $c_1$ as shorthand for $c(\boldsymbol{\theta}_1)$, $c_2$ for $c(\boldsymbol{\theta}_2)$, $\hat{c}_1$ for $\hat{c}_N(\boldsymbol{\theta}_1)$, and $\hat{c}_2$ for $\hat{c}_N(\boldsymbol{\theta}_2)$. Define $m = \frac{1}{2} \cdot (c_2 + c_1)$ as the midpoint between $c_1$ and $c_2$, and $\epsilon = c_2 - m = m - c_1 > 0$ as its distance from each of the two points. Since $\hat{c}_N$ is a consistent estimator for $c$, the estimate $\hat{c}_1$ comes arbitrarily close to the real cost $c_1$. That is, $\lim_{N \to \infty} P(|\hat{c}_1 - c_1| < \epsilon) = 1$. Since $|m - c_1| = \epsilon$, the estimate $\hat{c}_1$ cannot be greater than or equal to $m$: $\lim_{N \to \infty} P(\hat{c}_1 \geq m) = 0$.
Similarly, $\lim_{N \to \infty} P(\hat{c}_2 < m) = 0$. Since

$$
\begin{aligned}
P(\hat{c}_1 \geq \hat{c}_2) &= P(\hat{c}_1 \geq \hat{c}_2 \wedge \hat{c}_1 \geq m) + P(\hat{c}_1 \geq \hat{c}_2 \wedge \hat{c}_1 < m) \\
&= P(\hat{c}_1 \geq \hat{c}_2 \wedge \hat{c}_1 \geq m) + P(\hat{c}_1 \geq \hat{c}_2 \wedge \hat{c}_1 < m \wedge \hat{c}_2 < m) \\
&\leq P(\hat{c}_1 \geq m) + P(\hat{c}_2 < m),
\end{aligned}
$$

we have $\lim_{N \to \infty} P(\hat{c}_1 \geq \hat{c}_2) \leq \lim_{N \to \infty} (P(\hat{c}_1 \geq m) + P(\hat{c}_2 < m)) = 0 + 0 = 0$. □

Combining our two lemmata we can now show that in the limit, FocusedILS is guaranteed to converge to the true best parameter configuration.

**Theorem 9** (Convergence of FocusedILS). *When FocusedILS optimizes a cost statistic $c$ based on a consistent estimator $\hat{c}_N$, the probability that it finds the true optimal parameter configuration $\boldsymbol{\theta}^*$ approaches one as the number of ILS iterations goes to infinity.*

*Proof.* According to Lemma 6, $N(\boldsymbol{\theta})$ grows unboundedly for each $\boldsymbol{\theta} \in \Theta$. For each $\boldsymbol{\theta}_1$, $\boldsymbol{\theta}_2$, as $N(\boldsymbol{\theta}_1)$ and $N(\boldsymbol{\theta}_2)$ go to infinity, Lemma 8 states that in a pairwise comparison, the truly better configuration will be preferred. Thus eventually, FocusedILS visits all finitely many parameter configurations and prefers the best one over all others with probability arbitrarily close to one. □

We note that in many practical scenarios cost estimators may not be consistent—that is, they may fail to closely approximate the true performance of a given parameter configuration even for a large number of runs of the target algorithm. For example, when a finite training set, $\Pi$, is used during configuration rather than a distribution over problem instances, $\mathcal{D}$, then even for large $N$, $\hat{c}_N$ will only accurately reflect the cost of parameter configurations on the training set, $\Pi$. For small training sets, $\Pi$, the cost estimate based on $\Pi$ may differ substantially from the true cost as defined by performance across the entire distribution, $\mathcal{D}$. The larger the training set, $\Pi$, the smaller the expected difference (it vanishes as training set size goes to infinity). Thus, it is important to use large training sets (which are representative of the distribution of interest) whenever possible.

## 4. Adaptive Capping of Algorithm Runs

Now we consider the last of our dimensions of automated algorithm configuration, the cutoff time for each run of the target algorithm. We introduce an effective and simple capping technique that adaptively determines the cutoff time for each run. The motivation for this capping technique comes from a problem encountered by all configuration procedures considered in this article: often the search for a performance-optimizing parameter setting spends a lot of time with evaluating a parameter configuration that is much worse than other, previously-seen configurations.

Consider, for example, a case where parameter configuration $\boldsymbol{\theta}_1$ takes a total of 10 seconds to solve $N = 100$ instances (i.e., it has a mean runtime of 0.1 seconds per instance), and another parameter configuration $\boldsymbol{\theta}_2$ takes 100 seconds to solve the first of these instances. In order to compare the mean runtimes of $\boldsymbol{\theta}_1$ and $\boldsymbol{\theta}_2$ based on this set of instances, knowing all runtimes for $\boldsymbol{\theta}_1$, it is not necessary to run $\boldsymbol{\theta}_2$ on all 100 instances. Instead, we can already terminate the first run of $\boldsymbol{\theta}_2$ after $10 + \epsilon$ seconds. This results in a lower bound on $\boldsymbol{\theta}_2$'s mean runtime of $0.1 + \epsilon/100$ since the remaining 99 instances could take no less than zero time. This lower bound exceeds the mean runtime of $\boldsymbol{\theta}_1$, and so we can already be certain that the comparison will favour $\boldsymbol{\theta}_1$. This insight provides the basis for our *adaptive capping* technique.

## 4.1 Adaptive Capping in BasicILS

In this section, we introduce adaptive capping for BasicILS. We first introduce a trajectory-preserving version of adaptive capping (**TP capping**) that provably does not change BasicILS's search trajectory and can lead to large computational savings. We then modify this strategy heuristically to perform more aggressive adaptive capping (**Aggr capping**), potentially yielding even better performance in practice.

### 4.1.1 TRAJECTORY-PRESERVING CAPPING

Observe that all comparisons between parameter configurations in ParamILS are *pairwise*. In BasicILS($N$), these comparisons are based on Procedure $better_N(\theta_1, \theta_2)$, where $\theta_2$ is either the best configuration encountered in this ILS iteration or the best configuration of the last ILS iteration. Without adaptive capping, these comparisons can take a long time, since a poor parameter configuration $\theta$ can easily take more than an order of magnitude longer than good configurations.

For the case of optimizing the mean of non-negative cost functions (such as runtime or solution cost), we implement a bounded evaluation of a parameter configuration $\theta$ based on $N$ runs and a given performance bound in Procedure *objective* (see Procedure 4). This procedure sequentially performs runs for parameter configuration $\theta$ and after each run computes a lower bound on $\hat{c}_N(\theta)$ based on the $i \leq N$ runs performed so far. Specifically, for our objective of mean runtime we sum the runtimes of each of the $i$ runs, and divide this sum by $N$; since all runtimes must be nonnegative, this quantity lower bounds $\hat{c}_N(\theta)$. Once the lower bound exceeds the bound passed as an argument, we can skip the remaining runs for $\theta$. In order to pass the appropriate bounds to Procedure *objective*, we need to slightly modify Procedure $better_N$ (see Procedure 2 on page 274) for adaptive capping. Procedure *objective* now has a bound as an additional third argument, which is set to $\infty$ in line 1 of $better_N$, and to $\hat{c}_N(\theta_2)$ in line 2.

Because this approach results in the computation of exactly the same function $better_N$ as used in the original version of BasicILS, the modified procedure follows exactly the same search trajectory it would have followed without capping, but typically requires much less runtime. Hence, within the same amount of overall running time, this new version of BasicILS tends to be able to search a larger part of the parameter configuration space. Although in this work we focus on the objective of minimizing mean runtime for decision algorithms, we note that our adaptive capping approach can be applied easily to other configuration objectives.

### 4.1.2 AGGRESSIVE CAPPING

As we demonstrate in Section 6.4, the use of trajectory-preserving adaptive capping can result in substantial speedups of BasicILS. However, sometimes this approach is still less efficient than it could be. This is because the upper bound on cumulative runtime used for capping is computed from the best configuration encountered *in the current ILS iteration* (where a new ILS iteration begins after each perturbation), as opposed to the overall incumbent. After a perturbation has resulted in a new parameter configuration $\theta$, the new iteration's best configuration is initialized to $\theta$. In the frequent case that this new $\theta$ performs poorly, the capping criterion does not apply as quickly as when the comparison is performed against the overall incumbent.

To counteract this effect, we introduce a more aggressive capping strategy that can terminate the evaluation of a poorly-performing configuration at any time. In this heuristic extension of our adaptive capping technique, we bound the evaluation of *any* parameter configuration by the per-

---

**Procedure 4**: $objective(\boldsymbol{\theta}, N,$ optional parameter $bound)$

Procedure that computes $\hat{c}_N(\boldsymbol{\theta})$, either by performing new runs or by exploiting previous cached runs. An optional third parameter specifies a bound on the computation to be performed; when this parameter is not specified, the bound is taken to be $\infty$. For each $\boldsymbol{\theta}$, $N(\boldsymbol{\theta})$ is the number of runs performed for $\boldsymbol{\theta}$, i.e., the length of the global array $\mathbf{R}_{\boldsymbol{\theta}}$. When computing runtimes, we count unsuccessful runs as 10 times their cutoff time.

| | |
|---|---|
| **Input** | : Parameter configuration $\boldsymbol{\theta}$, number of runs, $N$, optional bound $bound$ |
| **Output** | : $\hat{c}_N(\boldsymbol{\theta})$ if $\hat{c}_N(\boldsymbol{\theta}) \leq bound$, otherwise a large constant (maxPossibleObjective) plus the number of instances that remain unsolved when the bound was exceeded |
| **Side Effect** | : Adds runs to the global cache of performed algorithm runs, $\mathbf{R}_{\boldsymbol{\theta}}$; updates global incumbent, $\boldsymbol{\theta}_{inc}$ |

*// ===== Maintain invariant: $N(\boldsymbol{\theta}_{inc}) \geq N(\boldsymbol{\theta})$ for any $\boldsymbol{\theta}$*

**1** **if** $\boldsymbol{\theta} \neq \boldsymbol{\theta}_{inc}$ **and** $N(\boldsymbol{\theta}_{inc}) < N$ **then**

**2**     $\hat{c}_N(\boldsymbol{\theta}_{inc}) \leftarrow objective(\boldsymbol{\theta}_{inc}, N, \infty)$    *// Adds $N - N(\boldsymbol{\theta}_{inc})$ runs to $\mathbf{R}_{\boldsymbol{\theta}_{inc}}$*

    *// ===== For aggressive capping, update bound.*

**3** **if** *Aggressive capping* **then** $bound \leftarrow \min(bound, bm \cdot \hat{c}_N(\boldsymbol{\theta}_{inc}))$

    *// ===== Update the run results in tuple $\mathbf{R}_{\boldsymbol{\theta}}$.*

**4** **for** $i = 1...N$ **do**

**5**     sum_runtime $\leftarrow$ sum of runtimes in $\mathbf{R}_{\boldsymbol{\theta}}[1], \ldots, \mathbf{R}_{\boldsymbol{\theta}}[i-1]$ *// Tuple indices starting at 1.*

**6**     $\kappa'_i \leftarrow \max(\kappa_{max}, N \cdot bound - $ sum_runtime$)$

**7**     **if** $N(\boldsymbol{\theta}) \geq i$ **then** $(\boldsymbol{\theta}, \pi_i, \kappa_i, o_i) \leftarrow \mathbf{R}_{\boldsymbol{\theta}}[i]$

**8**     **if** $N(\boldsymbol{\theta}) \geq i$ **and** $((\kappa_i \geq \kappa'_i$ **and** $o_i = $ *"unsuccessful"*$)$ **or** $(\kappa_i < \kappa'_i$ **and** $o_i \neq $ *"unsuccessful"*$))$

    **then** $o'_i \leftarrow o_i$ *// Previous run is longer yet unsuccessful or shorter yet successful $\Rightarrow$ can re-use result*

**9**     **else**

**10**       $o'_i \leftarrow$ objective from a newly executed run of $\mathcal{A}(\boldsymbol{\theta})$ on instance $\pi_i$ with seed $s_i$ and captime $\kappa_i$

**11**     $\mathbf{R}_{\boldsymbol{\theta}}[i] \leftarrow (\boldsymbol{\theta}, \pi_i, \kappa'_i, o'_i)$

**12**     **if** $1/N \cdot ($sum_runtime $+ o'_i) > bound$ **then return** maxPossibleObjective $+ (N+1) - i$

**13** **if** $N = N(\boldsymbol{\theta}_{inc})$ **and** (sum of runtimes in $\mathbf{R}_{\boldsymbol{\theta}}$) < (sum of runtimes in $\mathbf{R}_{\boldsymbol{\theta}_{inc}}$) **then** $\boldsymbol{\theta}_{inc} \leftarrow \boldsymbol{\theta}$

**14** **return** $1/N \cdot$ (sum of runtimes in $\mathbf{R}_{\boldsymbol{\theta}}$)

---

formance of the incumbent parameter configuration multiplied by a factor that we call the *bound multiplier*, $bm$. When a comparison between any two parameter configurations $\boldsymbol{\theta}$ and $\boldsymbol{\theta}'$ is performed and the evaluations of both are terminated preemptively, the configuration having solved more instances within the allowed time is taken to be the better one. (This behaviour is achieved by line 12 in Procedure $objective$, which keeps track of the number of instances solved when exceeding the bound.) Ties are broken to favour moving to a new parameter configuration instead of staying with the current one.

Depending on the bound multiplier, the use of this aggressive capping mechanism may change the search trajectory of BasicILS. For $bm = \infty$ the heuristic method reduces to our trajectory-preserving method, while a very aggressive setting of $bm = 1$ means that once we know a parameter configuration to be worse than the incumbent, we stop its evaluation. In our experiments we set $bm = 2$, meaning that once the lower bound on the performance of a configuration exceeds twice the performance of the incumbent solution, its evaluation is terminated. (In Section 8.4, we revisit this choice of $bm = 2$, configuring the parameters of ParamILS itself.)

## 4.2 Adaptive Capping in FocusedILS

The main difference between BasicILS and FocusedILS is that the latter adaptively varies the number of runs used to evaluate each parameter configuration. This difference complicates, but does not prevent the use of adaptive capping. This is because FocusedILS always compares pairs of parameter configurations based on the same number of runs for each configuration, even though this number can differ from one comparison to the next.

Thus, we can extend adaptive capping to FocusedILS by using separate bounds for every number of runs, $N$. Recall that FocusedILS never moves from one configuration, $\theta$, to a neighbouring configuration, $\theta'$, without performing at least as many runs for $\theta'$ as have been performed for $\theta$. Since we keep track of the performance of $\theta$ with any number of runs $M \leq N(\theta)$, a bound for the evaluation of $\theta'$ is always available. Therefore, we can implement both trajectory-preserving and aggressive capping as we did for BasicILS.

As for BasicILS, for FocusedILS the inner workings of adaptive capping are implemented in Procedure *objective* (see Procedure 4). We only need to modify Procedure *better$_{Foc}$* (see Procedure 3 on page 276) to call *objective* with the right bounds. This leads to the following changes in Procedure *better$_{Foc}$*. Subprocedure *dominates* on line 16 now takes a bound as an additional argument and passes it on to the two calls to *objective* in line 18. The two calls of *dominates* in line 10 and the one call in line 11 all use the bound $\hat{c}_{\theta_{max}}$. The three direct calls to *objective* in lines 8, 9, and 12 use bounds $\infty$, $\hat{c}_{\theta_{max}}$, and $\infty$, respectively.

## 5. Experimental Preliminaries

In this section we give background information about the computational experiments presented in the following sections. First, we describe the design of our experiments. Next, we present the configuration scenarios (algorithm/benchmark data combinations) studied in the following section. Finally, we describe the low-level details of our experimental setup.

## 5.1 Experimental Design

Here we describe our objective function and the methods we used for selecting instances and seeds.

### 5.1.1 CONFIGURATION OBJECTIVE: PENALIZED AVERAGE RUNTIME

In Section 2, we mentioned that algorithm configuration problems arise in the context of various different cost statistics. Indeed, in our past work we explored several of them: maximizing solution quality achieved in a given time, minimizing the runtime required to reach a given solution quality, and minimizing the runtime required to solve a single problem instance (Hutter et al., 2007).

In this work we focus on the objective of minimizing the mean runtime over instances from a distribution $\mathcal{D}$. This optimization objective naturally occurs in many practical applications. It also implies a strong correlation between $c(\theta)$ and the amount of time required to obtain a good empirical estimate of $c(\theta)$. This correlation helps to make our adaptive capping scheme effective.

One might wonder whether means are the right way to aggregate runtimes. In some preliminary experiments, we found that minimizing mean runtime led to parameter configurations with overall good runtime performance, including rather competitive median runtimes, while minimizing median runtime yielded less robust parameter configurations that timed out on a large (but $< 50\%$) fraction of the benchmark instances. However, when we encounter runs that do not terminate within the given cutoff time the mean is ill-defined. In order to penalize timeouts, we define the *penalized average runtime (PAR)* of a set of runs with cutoff time $\kappa_{max}$ to be the mean runtime over those runs, where unsuccessful runs are counted as $p \cdot \kappa_{max}$ with penalization constant $p \geq 1$. In this study, we use $p = 10$.

### 5.1.2 SELECTING INSTANCES AND SEEDS

As mentioned previously, often only a finite set $\Pi$ of instances is available upon which to evaluate our algorithm. This is the case in the experiments we report here. Throughout our study, all configuration experiments are performed on a training set containing half of the given benchmark instances. The remaining instances are solely used as a test set to evaluate the found parameter configurations.

For evaluations within ParamILS that are based on $N$ runs, we selected the $N$ instances and random number seeds to be used by following a common blocking technique (see, e.g., Birattari et al., 2002; Ridge & Kudenko, 2006). We ensured that whenever two parameter configurations were compared, their cost estimates were based on exactly the same instances and seeds. This serves to avoid noise effects due to differences between instances and the use of different seeds. For example, it prevents us from making the mistake of considering configuration $\theta$ to be better than configuration $\theta'$ just because $\theta$ was tested on easier instances.

When dealing with randomized target algorithms, there is also a tradeoff between the number of problem instances used and the number of independent runs performed on each instance. In the extreme case, for a given sample size $N$, one could perform $N$ runs on a single instance or a single run on $N$ different instances. This latter strategy is known to result in minimal variance of the estimator for common optimization objectives such as minimization of mean runtime (which we consider in this study) or maximization of mean solution quality (see, e.g., Birattari, 2004). Consequently, we only performed multiple runs per instance when we wanted to acquire more samples of the cost distribution than there were instances in the training set.

Based on these considerations, the configuration procedures we study in this article have been implemented to take a list of ⟨instance, random number seed⟩ pairs as one of their inputs. Empirical estimates $\hat{c}_N(\theta)$ of the cost statistic $c(\theta)$ to be optimized were determined from the first $N$ ⟨instance, seed⟩ pairs in that list. Each list of ⟨instance, seed⟩ pairs was constructed as follows. Given a training set consisting of $M$ problem instances, for $N \leq M$, we drew a sample of $N$ instances uniformly at random and without replacement and added them to the list. If we wished to evaluate an algorithm on more samples than we had training instances, which could happen in the case of randomized algorithms, we repeatedly drew random samples of size $M$ as described before, where each such batch corresponded to a random permutation of the $N$ training instances, and added a final sample of size $N \bmod M < M$, as in the case $N \leq M$. As each sample was drawn, it was paired with a random number seed that was chosen uniformly at random from the set of all possible seeds and added to the list of ⟨instance, seed⟩ pairs.

### 5.1.3 COMPARISON OF CONFIGURATION PROCEDURES

Since the choice of instances (and to some degree of seeds) is very important for the final outcome of the optimization, in our experimental evaluations we always performed a number of independent runs of each configuration procedure (typically 25). We created a separate list of instances and seeds for each run as explained above, where the $k$th run of each configuration procedure uses the same $k$th list of instances and seeds. (Note, however, that the disjoint test set used to measure performance of parameter configurations was identical for all runs.)

| Configuration scenario | Type of benchmark instances & citation |
|---|---|
| SAPS-SWGCP | Graph colouring (Gent, Hoos, Prosser & Walsh, 1999) |
| SPEAR-SWGCP | Graph colouring (Gent, Hoos, Prosser & Walsh, 1999) |
| SAPS-QCP | Quasigroup completion (Gomes & Selman, 1997) |
| SPEAR-QCP | Quasigroup completion (Gomes & Selman, 1997) |
| CPLEX-REGIONS100 | Combinatorial Auctions (CATS) (Leyton-Brown, Pearson & Shoham, 2000) |

Table 1: Overview of our five BROAD configuration scenarios.

| Algorithm | Parameter type | # parameters of type | # values considered | Total # configurations, $|\Theta|$ |
|---|---|---|---|---|
| SAPS | Continuous | 4 | 7 | $2\,401$ |
| SPEAR | Categorical | 10 | 2–20 | $8.34 \cdot 10^{17}$ |
| | Integer | 4 | 5–8 | |
| | Continuous | 12 | 3–6 | |
| CPLEX | Categorical | 50 | 2–7 | $1.38 \cdot 10^{37}$ |
| | Integer | 8 | 5–7 | |
| | Continuous | 5 | 3–5 | |

Table 2: Parameter overview for the algorithms we consider. More information on the parameters for each algorithm is given in the text. A detailed list of all parameters and the values we considered can be found in an online appendix at http://www.cs.ubc.ca/labs/beta/Projects/ParamILS/algorithms.html.

We performed a paired statistical test to compare the final results obtained in the runs of two configuration procedures. A paired test was required since the $k$th run of both procedures shared the same $k$th list of instances and seeds. In particular, we performed a two-sided paired Max-Wilcoxon test with the null hypothesis that there was no difference in the performances, considering $p$-values below 0.05 to be statistically significant. The $p$-values reported in all tables were derived using this test; $p$-values shown in parentheses refer to cases where the procedure we expected to perform better actually performed worse.

## 5.2 Configuration Scenarios

In Section 6, we analyze our configurators based on five configuration scenarios, each combining a high-performance algorithm with a widely-studied benchmark dataset. Table 1 gives an overview of these, which we dub the BROAD scenarios. The algorithms and benchmark instance sets used in these scenarios are described in detail in Sections 5.2.1 and 5.2.2, respectively. In these five BROAD configuration scenarios, we set fairly aggressive cutoff times of five seconds per run of the target algorithm and allowed each configuration procedure to execute the target algorithm for an aggregate runtime of five CPU hours. These short cutoff times and fairly short times for algorithm configuration were deliberately chosen to facilitate many configuration runs for each BROAD scenario. In contrast, in a second set of configuration scenarios (exclusively focusing on CPLEX), we set much larger cutoff times and allowed more time for configuration. We defer a description of these scenarios to Section 7.

### 5.2.1 TARGET ALGORITHMS

Our three target algorithms are listed in Table 2 along with their configurable parameters.

**Saps**    The first target algorithm used in our experiments was Saps, a high-performance dynamic local search algorithm for SAT solving (Hutter, Tompkins & Hoos, 2002) as implemented in UBC-SAT (Tompkins & Hoos, 2004). When introduced in 2002, Saps was a state-of-the-art solver, and it still performs competitively on many instances. We chose to study this algorithm because it is well known, it has relatively few parameters, and we are intimately familiar with it. Saps's four continuous parameters control the scaling and smoothing of clause weights, as well as the probability of random walk steps. The original default parameters were set manually based on experiments with prominent benchmark instances; this manual experimentation kept the percentage of random steps fixed and took up about one week of development time. Having subsequently gained more experience with Saps's parameters for more general problem classes (Hutter, Hamadi, Hoos & Leyton-Brown, 2006), we chose promising intervals for each parameter, including, but not centered at, the original default. We then picked seven possible values for each parameter spread uniformly across its respective interval, resulting in 2401 possible parameter configurations (these are exactly the same values as used by Hutter et al., 2007). As the starting configuration for ParamILS, we used the center point of each parameter's domain.

**Spear**    The second target algorithm we considered was Spear, a recent tree search algorithm for solving SAT problems. Spear is a state-of-the-art SAT solver for industrial instances, and with appropriate parameter settings it is the best available solver for certain types of hardware and software verification instances (Hutter, Babić, Hoos & Hu, 2007). Furthermore, configured with ParamILS, Spear won the quantifier-free bit-vector arithmetic category of the 2007 Satisfiability Modulo Theories Competition. Spear has 26 parameters, including ten categorical, four integer, and twelve continuous parameters, and their default values were manually engineered by its developer. (Manual tuning required about one week.) The categorical parameters mainly control heuristics for variable and value selection, clause sorting, resolution ordering, and enable or disable optimizations, such as the pure literal rule. The continuous and integer parameters mainly deal with activity, decay, and elimination of variables and clauses, as well as with the interval of randomized restarts and percentage of random choices. We discretized the integer and continuous parameters by choosing lower and upper bounds at reasonable values and allowing between three and eight discrete values spread relatively uniformly across the resulting interval, including the default, which served as the starting configuration for ParamILS. The number of discrete values was chosen according to our intuition about the importance of each parameter. After this discretization, there were $3.7 \cdot 10^{18}$ possible parameter configurations. Exploiting the fact that nine of the parameters are conditional (i.e., only relevant when other parameters take certain values) reduced this to $8.34 \cdot 10^{17}$ configurations.

**Cplex**    The third target algorithm we used was the commercial optimization tool Cplex 10.1.1, a massively parameterized algorithm for solving mixed integer programming (MIP) problems. Out of its 159 user-specifiable parameters, we identified 81 parameters that affect Cplex's search trajectory. We were careful to omit all parameters that change the problem formulation (e.g., by changing the numerical accuracy of a solution). Many Cplex parameters deal with MIP strategy heuristics (such as variable and branching heuristics, probing, dive type and subalgorithms) and with the amount and type of preprocessing to be performed. There are also nine parameters governing how frequently a different type of cut should be used (those parameters have up to four allowable magnitude values and the value "choose automatically"; note that this last value prevents the parameters from being ordinal). A considerable number of other parameters deal with simplex and barrier optimization, and with various other algorithm components. For categorical parameters with an automatic option, we considered all categorical values as well as the automatic one. In contrast, for continuous and integer parameters with an automatic option, we chose that option instead of hypothesizing values that might work well. We also identified some numerical parameters that primarily deal with numerical issues, and fixed those to their default values. For other numerical parameters, we chose up to five possible values that seemed sensible, including the default. For the many categorical parameters with an automatic option, we included the automatic option as a choice for the parameter, but also included all the manual options. Finally, we ended up with 63 configurable parameters, leading to $1.78 \cdot 10^{38}$ possible configurations. Exploiting the fact that seven of the CPLEX parameters were only relevant conditional on other parameters taking certain values, we reduced this to $1.38 \cdot 10^{37}$ distinct configurations. As the starting configuration for our configuration procedures, we used the default settings, which have been obtained by careful manual configuration on a broad range of MIP instances.

### 5.2.2 Benchmark Instances

We applied our target algorithms to three sets of benchmark instances: SAT-encoded quasi-group completion problems, SAT-encoded graph-colouring problems based on small world graphs, and MIP-encoded winner determination problems for combinatorial auctions. Each set consisted of 2000 instances, partitioned evenly into training and test sets.

**QCP**  Our first benchmark set contained $23\,000$ instances of the quasi-group completion problem (QCP), which has been widely studied by AI researchers. We generated these QCP instances around the solubility phase transition, using the parameters given by Gomes and Selman (1997). Specifically, the order $n$ was drawn uniformly from the interval $[26, 43]$, and the number of holes $H$ (open entries in the Latin square) was drawn uniformly from $[1.75, 2.3] \cdot n^{1.55}$. The resulting QCP instances were converted into SAT CNF format. For use with the complete solver, Spear, we sampled 2000 of these SAT instances uniformly at random. These had on average 1497 variables (standard deviation: 1094) and $13\,331$ clauses (standard deviation: $12\,473$), and 1182 of them were satisfiable. For use with the incomplete solver, Saps, we randomly sampled 2000 instances from the subset of satisfiable instances (determined using a complete algorithm); their number of variables and clauses were very similar to those used with Spear.

**SW-GCP**  Our second benchmark set contained $20\,000$ instances of the graph colouring problem (GCP) based on the small world (SW) graphs of Gent et al. (1999). Of these, we sampled 2000 instances uniformly at random for use with Spear; these had on average 1813 variables (standard deviation: 703) and $13\,902$ clauses (standard deviation: 5393), and 1109 of them were satisfiable. For use with Saps, we randomly sampled 2000 satisfiable instances (again, determined using a complete SAT algorithm), whose number of variables and clauses were very similar to those used with Spear.

**Regions100**  For our third benchmark set we generated 2000 instances of the combinatorial auction winner determination problem, encoded as mixed-integer linear programs (MILPs). We used the `regions` generator from the Combinatorial Auction Test Suite (Leyton-Brown et al., 2000), with the *goods* parameter set to 100 and the *bids* parameter set to 500. The resulting MILP instances contained 501 variables and 193 inequalities on average, with a standard deviation of 1.7 variables and 2.5 inequalities.

| Scenario | Test performance (penalized average runtime, in CPU seconds) | | | | | Fig. |
|---|---|---|---|---|---|---|
| | | mean ± stddev. for 10 runs | | Run with best training performance | | |
| | Default | BasicILS | FocusedILS | BasicILS | FocusedILS | |
| SAPS-SWGCP | 20.41 | 0.32 ± 0.06 | **0.32 ± 0.05** | **0.26** | **0.26** | 2(a) |
| SPEAR-SWGCP | 9.74 | **8.05 ± 0.9** | 8.3 ± 1.1 | 6.8 | **6.6** | 2(b) |
| SAPS-QCP | 12.97 | 4.86 ± 0.56 | **4.70 ± 0.39** | 4.85 | **4.29** | 2(c) |
| SPEAR-QCP | 2.65 | 1.39 ± 0.33 | **1.29 ± 0.2** | **1.16** | 1.21 | 2(d) |
| CPLEX-REGIONS100 | 1.61 | 0.5 ± 0.3 | **0.35 ± 0.04** | 0.35 | **0.32** | 2(e) |

Table 3: Performance comparison of the default parameter configuration and the configurations found with BasicILS and FocusedILS (both with Aggr Capping and $bm = 2$). For each configuration scenario, we list test performance (penalized average runtime over 1000 test instances, in CPU seconds) of the algorithm default, mean ± stddev of test performance across 25 runs of BasicILS(100) & FocusedILS (run for five CPU hours each), and the test performance of the run of BasicILS and FocusedILS that was best in terms of *training* performance. Boldface indicates the better of BasicILS and FocusedILS. The algorithm configurations found in FocusedILS's run with the best training performance are listed in an online appendix at `http://www.cs.ubc.ca/labs/beta/Projects/ParamILS/results.html`. Column "Fig." gives a reference to a scatter plot comparing the performance of those configurations against the algorithm defaults.

## 5.3 Experimental Setup

We carried out all of our experiments on a cluster of 55 dual 3.2GHz Intel Xeon PCs with 2MB cache and 2GB RAM, running OpenSuSE Linux 10.1. We measured runtimes as CPU time on these reference machines. All our configuration procedures were implemented as Ruby scripts, and we do not include the runtime of these scripts in the configuration time. In "easy" configuration scenarios, where most algorithm runs finish in milliseconds, the overhead of our scripts can be substantial. Indeed, the longest configuration run we observed took 24 hours to execute five hours worth of target algorithm runtime. In contrast, for the harder CPLEX scenarios described in Section 7 we observed virtually no overhead.

## 6. Empirical Evaluation of BasicILS, FocusedILS and Adaptive Capping

In this section, we use our BROAD scenarios to empirically study the performance of BasicILS($N$) and FocusedILS, as well as the effect of adaptive capping. We first demonstrate the large speedups ParamILS achieved over the default parameters and then study the components responsible for this success.

### 6.1 Empirical Comparison of Default and Optimized Parameter Configurations

In this section, for each of our five BROAD configuration scenarios, we compare the performance of the respective algorithm's default parameter configuration against the final configurations found by BasicILS(100) and FocusedILS. Table 3 and especially Figure 2 show that the configurators led to very substantial speedups.

In Table 3, we report the final performance achieved by 25 independent runs of each configurator. For each independent configuration run, we used a different set of training instances and seeds (constructed as described in Section 5.1.2). We note that there was often a rather large variance in the performances found in different runs of the configurators, and that the configuration found

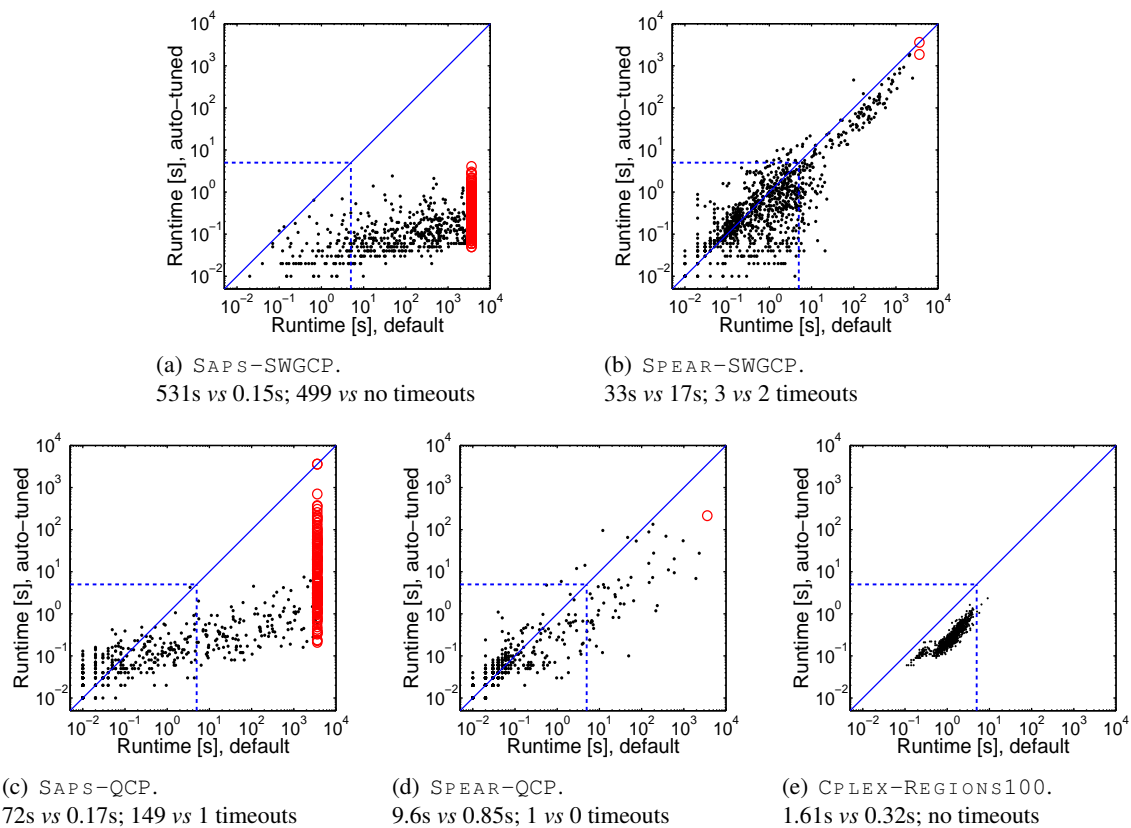

**(a)** SAPS-SWGCP.
531s *vs* 0.15s; 499 *vs* no timeouts

**(b)** SPEAR-SWGCP.
33s *vs* 17s; 3 *vs* 2 timeouts

**(c)** SAPS-QCP.
72s *vs* 0.17s; 149 *vs* 1 timeouts

**(d)** SPEAR-QCP.
9.6s *vs* 0.85s; 1 *vs* 0 timeouts

**(e)** CPLEX-REGIONS100.
1.61s *vs* 0.32s; no timeouts

Figure 2: Comparison of default *vs* automatically-determined parameter configurations for our five BROAD configuration scenarios. Each dot represents one test instance; timeouts (after one CPU hour) are denoted by circles. The dashed line at five CPU seconds indicates the cutoff time of the target algorithm used during the configuration process. The subfigure captions give mean runtimes for the instances solved by both of the configurations (default *vs* optimized), as well as the number of timeouts for each.

in the run with the best *training* performance also tended to yield better *test* performance than the others. For that reason, we used that configuration as the result of algorithm configuration. (Note that choosing the configuration found in the run with the best training set performance is a perfectly legitimate procedure since it does not require knowledge of the test set. Of course, the improvements thus achieved come at the price of increased overall running time, but the independent runs of the configurator can easily be performed in parallel.)

In Figure 2, we compare the performance of this automatically-found parameter configuration against the default configuration, when runs are allowed to last up to an hour. The speedups are more obvious in this figure than in Table 3, since the penalized average runtime in that table counts runtimes larger than five seconds as fifty seconds (ten times the cutoff of five seconds), whereas the data in the figure uses a much larger cutoff time. The larger speedups are most apparent for scenarios SAPS-SWGCP, SAPS-QCP, and SPEAR-QCP: their corresponding speedup factors in mean runtime are now 3540, 416 and 11, respectively (see Figure 2).

---

**Algorithm 5: RandomSearch**($N, \boldsymbol{\theta}_0$)

Outline of random search in parameter configuration space; $\boldsymbol{\theta}_{inc}$ denotes the incumbent parameter configuration, *better$_N$* compares two configurations based on the first $N$ instances from the training set.

---

| | |
|---|---|
| **Input** | : Number of runs to use for evaluating parameter configurations, $N$; initial configuration $\boldsymbol{\theta}_0 \in \Theta$. |
| **Output** | : Best parameter configuration $\boldsymbol{\theta}_{inc}$ found. |

**1** $\boldsymbol{\theta}_{inc} \leftarrow \boldsymbol{\theta}_0$;
**2 while not** *TerminationCriterion()* **do**
**3**     $\boldsymbol{\theta} \leftarrow$ random $\boldsymbol{\theta} \in \Theta$;
**4**     **if** better$_N$($\boldsymbol{\theta}, \boldsymbol{\theta}_{inc}$) **then**
**5**        $\boldsymbol{\theta}_{inc} \leftarrow \boldsymbol{\theta}$;

**6 return** $\boldsymbol{\theta}_{inc}$

---

## 6.2 Empirical Comparison of BasicILS and Simple Baselines

In this section, we evaluate the effectiveness of BasicILS($N$) against two of its components:

- a simple random search, used in BasicILS for initialization (we dub it RandomSearch($N$) and provide pseudocode for it in Algorithm 5); and

- a simple local search, the same type of iterative first improvement search used in BasicILS($N$) (we dub it SimpleLS($N$)).

To evaluate one component at a time, in this section and in Section 6.3 we study our algorithms without adaptive capping. We then investigate the effect of our adaptive capping methods in Section 6.4.

If there is sufficient structure in the search space, we expect BasicILS to outperform Random-Search. If there are local minima, we expect BasicILS to perform better than simple local search. Our experiments showed that BasicILS did indeed offer the best performance.

Here, we are solely interested in comparing how effectively the approaches search the space of parameter configurations (and not how the found parameter configurations generalize to unseen test instances). Thus, in order to reduce variance in our comparisons, we compare the configuration methods in terms of their performance on the training set.

In Table 4, we compare BasicILS against RandomSearch for our BROAD configuration scenarios. On average, BasicILS always performed better, and in three of the five scenarios, the difference was statistically significant as judged by a paired Max-Wilcoxon test (see Section 5.1.3). Table 4 also lists the performance of the default parameter configuration for each of the scenarios. We note that both BasicILS and RandomSearch consistently achieved substantial (and statistically significant) improvements over these default configurations.

Next, we compared BasicILS against its second component, SimpleLS. This basic local search is identical to BasicILS, but stops at the first local minimum encountered. We used it in order to study whether local minima pose a problem for simple first improvement search. Table 5 shows that in the three configuration scenarios where BasicILS had time to perform multiple ILS iterations, its training set performance was statistically significantly better than that of SimpleLS. Thus, we conclude that the search space contains structure that can be exploited with a local search algorithm as well as local minima that can limit the performance of iterative improvement search.

| Scenario | Training performance (penalized average runtime, in CPU seconds) | | | $p$-value |
|---|---|---|---|---|
| | Default | RandomSearch(100) | BasicILS(100) | |
| SAPS-SWGCP | 19.93 | $0.46 \pm 0.34$ | $\mathbf{0.38 \pm 0.19}$ | 0.94 |
| SPEAR-SWGCP | 10.61 | $7.02 \pm 1.11$ | $\mathbf{6.78 \pm 1.73}$ | 0.18 |
| SAPS-QCP | 12.71 | $3.96 \pm 1.185$ | $\mathbf{3.19 \pm 1.19}$ | $\mathbf{1.4 \cdot 10^{-5}}$ |
| SPEAR-QCP | 2.77 | $0.58 \pm 0.59$ | $\mathbf{0.36 \pm 0.39}$ | $\mathbf{0.007}$ |
| CPLEX-REGIONS100 | 1.61 | $1.45 \pm 0.35$ | $\mathbf{0.72 \pm 0.45}$ | $\mathbf{1.2 \cdot 10^{-5}}$ |

Table 4: Comparison of RandomSearch(100) and BasicILS(100), both without adaptive capping. The table shows training performance (penalized average runtime over $N = 100$ training instances, in CPU seconds). Note that both approaches yielded substantially better results than the default configuration, and that BasicILS performed statistically significantly better than RandomSearch in three of the five BROAD configuration scenarios as judged by a paired Max-Wilcoxon test (see Section 5.1.3).

| Scenario | SimpleLS(100) | BasicILS(100) | | $p$-value |
|---|---|---|---|---|
| | Performance | Performance | Avg. # ILS iterations | |
| SAPS-SWGCP | $0.5 \pm 0.39$ | $\mathbf{0.38 \pm 0.19}$ | 2.6 | $\mathbf{9.8 \cdot 10^{-4}}$ |
| SAPS-QCP | $3.60 \pm 1.39$ | $\mathbf{3.19 \pm 1.19}$ | 5.6 | $\mathbf{4.4 \cdot 10^{-4}}$ |
| SPEAR-QCP | $0.4 \pm 0.39$ | $\mathbf{0.36 \pm 0.39}$ | 1.64 | $\mathbf{0.008}$ |

Table 5: Comparison of SimpleLS(100) and BasicILS(100), both without adaptive capping. The table shows training performance (penalized average runtime over $N = 100$ training instances, in CPU seconds). In configuration scenarios SPEAR-SWGCP and CPLEX-REGIONS100, BasicILS did not complete its first ILS iteration in any of the 25 runs; the two approaches were thus identical and are not listed here. In all other configuration scenarios, BasicILS found significantly better configurations than SimpleLS.

## 6.3 Empirical Comparison of FocusedILS and BasicILS

In this section we investigate FocusedILS's performance experimentally. In contrast to our previous comparison of RandomSearch, SimpleLS, and BasicILS using *training* performance, we now compare FocusedILS against BasicILS using *test* performance. This is because—in contrast to BasicILS and SimpleLS—FocusedILS grows the number of target algorithm runs used to evaluate a parameter configuration over time. Even different runs of FocusedILS (using different training sets and random seeds) do not use the same number of target algorithm runs to evaluate parameter configurations. However, they all eventually aim to optimize the same cost statistic, $c$, and therefore test set performance (an unbiased estimator of $c$) provides a fairer basis for comparison than training performance. We only compare FocusedILS to BasicILS, since BasicILS already outperformed RandomSearch and SimpleLS in Section 6.2.

Figure 3 compares the test performance of FocusedILS and BasicILS($N$) with $N = 1$, 10 and 100. Using a single target algorithm run to evaluate each parameter configuration, BasicILS(1) was fast, but did not generalize well to the test set at all. For example, in configuration scenario SAPS-SWGCP, BasicILS(1) selected a parameter configuration whose test performance turned out to be even worse than the default. On the other hand, using a large number of target algorithm runs for each evaluation resulted in a very slow search, but eventually led to parameter configurations with good test performance. FocusedILS aims to achieve a fast search *and* good generalization to the test set. For the configuration scenarios in Figure 3, FocusedILS started quickly and also led to the best final performance.

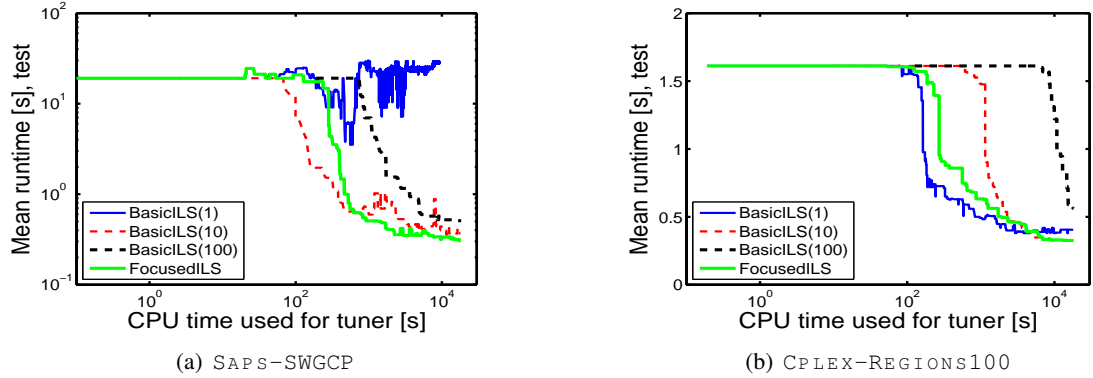

(a) SAPS-SWGCP

(b) CPLEX-REGIONS100

Figure 3: Comparison of BasicILS($N$) with $N = 1$, 10, and 100 *vs* FocusedILS, both without adaptive capping. We show the median of test performance (penalized average runtime across 1 000 test instances) across 25 runs of the configurators for two scenarios. Performance in the other three BROAD scenarios was qualitatively similar: BasicILS(1) was the fastest to move away from the starting parameter configuration, but its performance was not robust at all; BasicILS(10) was a rather good compromise between speed and generalization performance, but given enough time was outperformed by BasicILS(100). FocusedILS started finding good configurations quickly (except for scenario SPEAR-QCP, where it took even longer than BasicILS(100) to improve over the default) and always was amongst the best approaches at the end of the configuration process.

| | Test performance (penalized average runtime, in CPU seconds) | | | |
| Scenario | Default | BasicILS(100) | FocusedILS | *p*-value |
|---|---|---|---|---|
| SAPS-SWGCP | 20.41 | $0.59 \pm 0.28$ | $\mathbf{0.32 \pm 0.08}$ | $\mathbf{1.4 \cdot 10^{-4}}$ |
| SPEAR-SWGCP | 9.74 | $\mathbf{8.13 \pm 0.95}$ | $8.40 \pm 0.92$ | (0.21) |
| SAPS-QCP | 12.97 | $4.87 \pm 0.34$ | $\mathbf{4.69 \pm 0.40}$ | $\mathbf{0.042}$ |
| SPEAR-QCP | 2.65 | $\mathbf{1.32 \pm 0.34}$ | $1.35 \pm 0.20$ | (0.66) |
| CPLEX-REGIONS100 | 1.61 | $0.72 \pm 0.45$ | $\mathbf{0.33 \pm 0.03}$ | $\mathbf{1.2 \cdot 10^{-5}}$ |

Table 6: Comparison of BasicILS(100) and FocusedILS, both without adaptive capping. The table shows test performance (penalized average runtime over 1 000 test instances, in CPU seconds). For each configuration scenario, we report test performance of the default parameter configuration, mean $\pm$ stddev of the test performance reached by 25 runs of BasicILS(100) and FocusedILS, and the *p*-value for a paired Max-Wilcoxon test (see Section 5.1.3) for the difference of the two configurator's performance.

We compare the performance of FocusedILS and BasicILS(100) for all configuration scenarios in Table 6. For three SAPS and CPLEX scenarios, FocusedILS performed statistically significantly better than BasicILS(100). These results are consistent with our past work in which FocusedILS achieved statistically significantly better performance than BasicILS(100) (Hutter et al., 2007). However, we found that in both configuration scenarios involving the SPEAR algorithm, BasicILS(100) actually performed better on average than FocusedILS, albeit not statistically significantly. We attribute this to the fact that for a complete, industrial solver such as SPEAR, the two benchmark distributions QCP and SWGCP are quite heterogeneous. We expect FocusedILS to have problems in dealing with highly heterogeneous distributions, due to the fact that it frequently tries to extrapolate performance based on a few runs per parameter configuration.

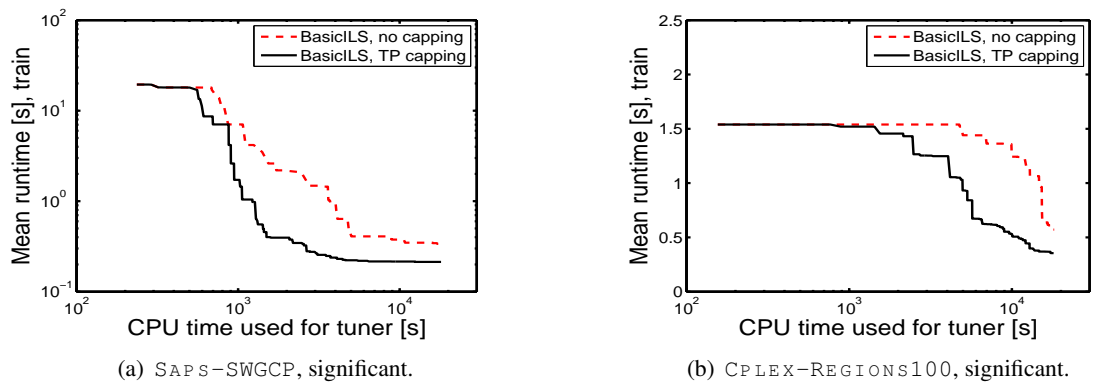

(a) Saps-SWGCP, significant.  (b) Cplex-Regions100, significant.

Figure 4: Speedup of BasicILS by adaptive capping for two configuration scenarios. We performed 25 runs of BasicILS(100) without adaptive capping and with TP capping. For each time step, we computed training performance of each run of the configurator (penalized average runtime over $N = 100$ training instances) and plot the median over the 25 runs.

| Scenario | Training performance (penalized average runtime) | | | Avg. # ILS iterations | |
|---|---|---|---|---|---|
| | No capping | TP capping | $p$-value | No capping | TP capping |
| Saps-SWGCP | $0.38 \pm 0.19$ | $\mathbf{0.24 \pm 0.05}$ | $\mathbf{6.1 \cdot 10^{-5}}$ | 3 | **12** |
| Spear-SWGCP | $6.78 \pm 1.73$ | $\mathbf{6.65 \pm 1.48}$ | **0.01** | 1 | 1 |
| Saps-QCP | $3.19 \pm 1.19$ | $\mathbf{2.96 \pm 1.13}$ | $\mathbf{9.8 \cdot 10^{-4}}$ | 6 | **10** |
| Spear-QCP | $0.361 \pm 0.39$ | $\mathbf{0.356 \pm 0.44}$ | 0.66 | 2 | **3** |
| Cplex-Regions100 | $0.67 \pm 0.35$ | $\mathbf{0.47 \pm 0.26}$ | $\mathbf{7.3 \cdot 10^{-4}}$ | 1 | 1 |

Table 7: Effect of adaptive capping for BasicILS(100). We show training performance (penalized average runtime on $N = 100$ training instances, in CPU seconds). For each configuration scenario, we report mean $\pm$ stddev of the final training performance reached by 25 runs of the configurator without capping and with TP capping, the $p$-value for a paired Max-Wilcoxon test for their difference (see Section 5.1.3), as well as the average number of ILS iterations performed by the respective configurator.

## 6.4 Empirical Evaluation of Adaptive Capping in BasicILS and FocusedILS

We now present experimental evidence that the use of adaptive capping has a strong impact on the performance of BasicILS and FocusedILS.

Figure 4 illustrates the extent to which TP capping sped up BasicILS for two configuration scenarios. In both cases, capping helped to improve training performance substantially; for Saps-SWGCP, BasicILS found the same solutions up to about an order of magnitude faster than without capping. Table 7 quantifies the speedups for all five Broad configuration scenarios. TP capping enabled up to four times as many ILS iterations (in Saps-SWGCP) and improved average performance in all scenarios. The improvement was statistically significant in all scenarios, except Spear-QCP.

Aggressive capping further improved BasicILS performance for one scenario. For scenario Saps-SWGCP, it increased the number of ILS iterations completed within the configuration time from 12 to 219, leading to a significant improvement in performance. In the first ILS iteration of BasicILS, both capping techniques are identical (the best configuration in that iteration is always the incumbent). Thus, we did not observe a difference on configuration scenarios Spear-SWGCP and Cplex-Regions100, for which none of the 25 runs of the configurator finished its first ILS iteration. For the remaining two configuration scenarios, the differences were insignificant.

| **Number of ILS iterations performed** | | | | | |
|---|---|---|---|---|---|
| Scenario | No capping | TP capping | $p$-value | Aggr capping | $p$-value |
| Saps-SWGCP | $121 \pm 12$ | $166 \pm 15$ | $\mathbf{1.2 \cdot 10^{-5}}$ | $\mathbf{244 \pm 19}$ | $\mathbf{1.2 \cdot 10^{-5}}$ |
| Spear-SWGCP | $37 \pm 12$ | $43 \pm 15$ | $\mathbf{0.0026}$ | $\mathbf{47 \pm 18}$ | $\mathbf{9 \cdot 10^{-5}}$ |
| Saps-QCP | $142 \pm 18$ | $143 \pm 22$ | $0.54$ | $\mathbf{156 \pm 28}$ | $\mathbf{0.016}$ |
| Spear-QCP | $153 \pm 49$ | $165 \pm 41$ | $\mathbf{0.03}$ | $\mathbf{213 \pm 62}$ | $\mathbf{1.2 \cdot 10^{-5}}$ |
| Cplex-Regions100 | $36 \pm 13$ | $40 \pm 16$ | $0.26$ | $\mathbf{54 \pm 15}$ | $\mathbf{1.8 \cdot 10^{-5}}$ |
| **Number of runs performed for the incumbent parameter configuration** | | | | | |
| Scenario | No capping | TP capping | $p$-value | Aggr capping | $p$-value |
| Saps-SWGCP | $993 \pm 211$ | $1258 \pm 262$ | $\mathbf{4.7 \cdot 10^{-4}}$ | $\mathbf{1818 \pm 243}$ | $\mathbf{1.2 \cdot 10^{-5}}$ |
| Spear-SWGCP | $503 \pm 265$ | $476 \pm 238$ | $(0.58)$ | $\mathbf{642 \pm 288}$ | $\mathbf{0.009}$ |
| Saps-QCP | $1575 \pm 385$ | $1701 \pm 318$ | $0.065$ | $\mathbf{1732 \pm 340}$ | $0.084$ |
| Spear-QCP | $836 \pm 509$ | $1130 \pm 557$ | $\mathbf{0.02}$ | $\mathbf{1215 \pm 501}$ | $\mathbf{0.003}$ |
| Cplex-Regions100 | $761 \pm 215$ | $795 \pm 184$ | $0.40$ | $\mathbf{866 \pm 232}$ | $0.07$ |

Table 8: Effect of adaptive capping on search progress in FocusedILS, as measured by the number of ILS iterations performed and the number of runs performed for the incumbent parameter configuration. For each configuration scenario, we report mean $\pm$ stddev of both of these measures across 25 runs of the configurator without capping, with TP capping, and with Aggr capping, as well as the $p$-values for paired Max-Wilcoxon tests (see Section 5.1.3) for the differences between no capping and TP capping; and between no capping and Aggr capping.

We now evaluate the usefulness of capping for FocusedILS. Training performance is not a useful quantity in the context of comparing different versions of FocusedILS, since the number of target algorithm runs this measure is based on varies widely between runs of the configurator. Instead, we used two other measures to quantify search progress: the number of ILS iterations performed and the number of target algorithm runs performed for the incumbent parameter configuration. Table 8 shows these two measures for our five `Broad` configuration scenarios and the three capping schemes (none, TP, Aggr). FocusedILS with TP capping achieved higher values than without capping for all scenarios and both measures (although only some of the differences were statistically significant). Aggressive capping increased both measures further for all scenarios, and most of the differences between no capping and aggressive capping were statistically significant. Figure 5 demonstrates that for two configuration scenarios FocusedILS with capping reached the same solution qualities more quickly than without capping. However, after finding the respective configurations, FocusedILS showed no further significant improvement.

Recall that the experiments in Section 6.2 and 6.3 compared our various configurators without adaptive capping. One might wonder how these comparisons change in the presence of adaptive capping. Indeed, adaptive capping also worked "out of the box" for RandomSearch and enabled it to evaluate between 3.4 and 33 times as many configurations than without capping. This improvement significantly improved the simple algorithm RandomSearch to the point where its average performance came within 1% of the one of BasicILS for two domains (`Saps-SWGCP` and `Spear-SWGCP`; compare the much larger differences without capping reported in Table 4). For `Spear-QCP`, there was still a 25% difference in average performance, but this result was not significant. Finally, for `Saps-QCP` and `Cplex-Regions100` the difference was still substantial and significant (22% and 55% difference in average performance, with p-values $5.2 \cdot 10^{-5}$ and $0.0013$, respectively).

Adaptive capping also reduced the gap between BasicILS and FocusedILS. In particular, for `Saps-SWGCP`, where, even without adaptive capping, FocusedILS achieved the best performance we have encountered for this scenario, BasicILS caught up when using adaptive capping. Similarly, for CPLEX-REGIONS100, FocusedILS already performed very well without adaptive capping while BasicILS did not. Here, BasicILS improved based on adaptive capping, but still could not rival FocusedILS. For the other scenarios, adaptive capping did not affect the relative performance much; compare Tables 6 (without capping) and 3 (with capping) for details.

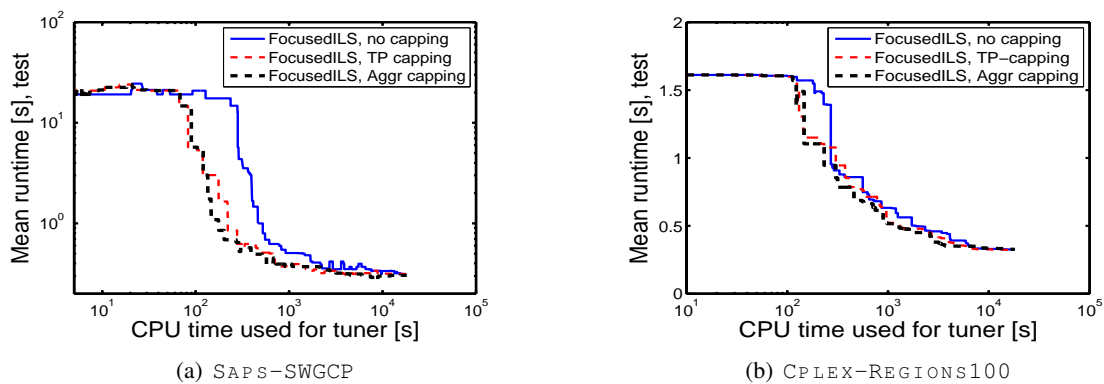

(a) SAPS-SWGCP        (b) CPLEX-REGIONS100

Figure 5: Speedup of FocusedILS by adaptive capping for two configuration scenarios. We performed 25 runs of FocusedILS without adaptive capping, with TP capping and with Aggr capping. For each time step, we computed the test performance of each run of the configurator (penalized average runtime over 1000 test instances) and plot the median over the 25 runs. The differences at the end of the trajectory were not statistically significant. However, with capping the time required to achieve that quality was lower in these two configuration scenarios. In the other three scenarios, the gains due to capping were smaller.

## 7. Case Study: Configuring CPLEX for Real-World Benchmarks

In this section, we demonstrate that ParamILS can improve the performance of the commercial optimization tool CPLEX for a variety of interesting benchmark distributions. To our best knowledge, this is the first published study on automatically configuring CPLEX.

We use five CPLEX configuration scenarios. For these, we collected a wide range of MIP benchmarks from public benchmark libraries and other researchers, and split each of them 50:50 into disjoint training and test sets; we detail them in the following.

- **Regions200** This set is almost identical to the **Regions100** set (described in Section 5.2.2 and used throughout the paper), but its instances are much larger. We generated 2 000 MILP instances with the generator provided with the Combinatorial Auction Test Suite (Leyton-Brown et al., 2000), based on the `regions` option with the goods parameter set to 200 and the bids parameter set to 1 000. These instances contain an average of 1 002 variables and 385 inequalities, with respective standard deviations of 1.7 and 3.4.

- **MJA** This set comprises 343 machine-job assignment instances encoded as mixed integer quadratically constrained programs (MIQCP). It was obtained from the Berkeley Computational Optimization Lab[5] and was introduced by Aktürk, Atamtürk and S. Gürel (2007). These instances contain an average of 2 769 variables and 2 255 constraints, with respective standard deviations of 2 133 and 1 592.

---

5. `http://www.ieor.berkeley.edu/~atamturk/bcol/`, where this set is called `conic.sch`

- **CLS** This set comprises 100 capacitated lot-sizing instances encoded as mixed integer linear programs (MILP). It was also obtained from the Berkeley Computational Optimization Lab and was introduced by Atamtürk and Muñoz (2004). All 100 instances contain 181 variables and 180 constraints.

- **MIK** This set of 120 MILP-encoded mixed-integer knapsack instances was also obtained from the Berkeley Computational Optimization Lab and was originally introduced by Atamtürk (2003). These instances contain an average of 384 variables and 151 constraints, with respective standard deviations of 309 and 127.

- **QP** This set of quadratic programs originated from RNA energy parameter optimization (Andronescu, Condon, Hoos, Mathews & Murphy, 2007). Mirela Andronescu generated 2 000 instances for our experiments. These instances contain $9\,366 \pm 7\,165$ variables and $9\,191 \pm 7\,186$ constraints. Since the instances are polynomial-time solvable quadratic programs, we set a large number of inconsequential CPLEX parameters concerning the branch and cut mechanism to their default values, ending up with 27 categorical, 2 integer and 2 continuous parameters to be configured, for a discretized parameter configuration space of size $3.27 \times 10^{17}$.

To study ParamILS's behavior for these harder problems, we set significantly longer cutoff times for these CPLEX scenarios than for the BROAD scenarios from the previous section. Specifically, we used a cutoff time of 300 CPU seconds for each run of the target algorithm during training, and allotted two CPU days for every run of each of the configurators. As always, our configuration objective was to minimize penalized average runtime with a penalization constant of 10.

In Table 9, we compare the performance of CPLEX's default parameter configuration with the final parameter configurations found by BasicILS(100) and FocusedILS (both with aggressive capping and $bm = 2$). Note that, similar to the situation described in Section 6.1, in some configuration scenarios (e.g., CPLEX-CLS, CPLEX-MIK) there was substantial variance between the different runs of the configurators, and the run with the best training performance yielded a parameter configuration that was also very good on the test set. While BasicILS outperformed FocusedILS in 3 of these 5 scenarios in terms of mean test performance across the ten runs, FocusedILS achieved the better test performance for the run with the best training performance for all but one scenario (in which it performed almost as well). For scenarios CPLEX-REGIONS200 and CPLEX-CLS, FocusedILS performed substantially better than BasicILS.

Note that in all CPLEX configuration scenarios we considered, both BasicILS and FocusedILS found parameter configurations that were better than the algorithm defaults, sometimes by over an order of magnitude. This is particularly noteworthy since ILOG expended substantial effort to determine strong default CPLEX parameters. In Figure 6, we provide scatter plots for all five scenarios. For CPLEX-REGIONS200, CPLEX-CONIC.SCH, CPLEX-CLS, and CPLEX-MIK, speedups were quite consistent across instances (with average speedup factors reaching from 2 for CPLEX-CONIC.SCH to 23 for CPLEX-MIK). Finally, for CPLEX-QP we see an interesting failure mode of ParamILS. The optimized parameter configuration achieved good performance with the cutoff time used for the configuration process (300 CPU seconds, see Figure 6(f)), but this performance did not carry over to the higher cutoff time we used in our tests (3600 CPU seconds, see Figure 6(e)). Thus, the parameter configuration found by FocusedILS *did* generalize well to previously unseen test data, but *not* to larger cutoff times.

---

6. For configuration scenario CPLEX-MIK, nine out of ten runs of FocusedILS yielded parameter configurations with average runtimes smaller than two seconds. One run, however, demonstrated an interesting failure mode of FocusedILS with aggressive capping. Capping too aggressively caused every CPLEX run to be unsuccessful, such that FocusedILS selected a configuration which did not manage to solve a single instance in the test set. Counting unsuccessful runs as ten times the cutoff time, this resulted in an average runtime of $10 \cdot 300 = 3000$ seconds for this run. (For full details, see Section 8.1 of Hutter, 2009).

| Scenario | Test performance (penalized average runtime, in CPU seconds) | | | | | Fig. |
|---|---|---|---|---|---|---|
| | Default | mean ± stddev. for 10 runs | | Run with best training performance | | |
| | | BasicILS | FocusedILS | BasicILS | FocusedILS | |
| CPLEX-REGIONS200 | 72 | $45 \pm 24$ | $\mathbf{11.4 \pm 0.9}$ | 15 | **10.5** | 6(a) |
| CPLEX-CONIC.SCH | 5.37 | $\mathbf{2.27 \pm 0.11}$ | $2.4 \pm 0.29$ | **2.14** | 2.35 | 6(b) |
| CPLEX-CLS | 712 | $443 \pm 294$ | $\mathbf{327 \pm 860}$ | 80 | **23.4** | 6(c) |
| CPLEX-MIK | 64.8 | $\mathbf{20 \pm 27}$ | $301 \pm 948^{\,6}$ | 1.72 | **1.19** | 6(d) |
| CPLEX-QP | 969 | $\mathbf{755 \pm 214}$ | $827 \pm 306$ | 528 | **525** | 6(e) |

Table 9: Experimental results for our CPLEX configuration scenarios. For each configuration scenario, we list test performance (penalized average runtime over test instances) of the algorithm default, mean ± stddev of test performance across ten runs of BasicILS(100) & FocusedILS (run for two CPU days each), and the test performance of the run of BasicILS and FocusedILS that is best in terms of *training* performance. Boldface indicates the better of BasicILS and FocusedILS. The algorithm configurations found in FocusedILS's run with the best training performance are listed in an online appendix at `http://www.cs.ubc.ca/labs/beta/Projects/ParamILS/results.html`. Column "Fig." gives a reference to a scatter plot comparing the performance of those configurations against the algorithm defaults.

# 8. Review of Other ParamILS Applications

In this section, we review a number of other applications of ParamILS—some of them dating back to earlier stages of its development, others very recent—that demonstrate its utility and versatility.

## 8.1 Configuration of SAPS, GLS$^+$ and SAT4J

Hutter et al. (2007), in the first publication on ParamILS, reported experiments on three target algorithms to demonstrate the effectiveness of the approach: the SAT algorithm SAPS (which has 4 numerical parameters), the local search algorithm GLS$^+$ for solving the most probable explanation (MPE) problem in Bayesian networks (which has 5 numerical parameters; Hutter, Hoos & Stützle, 2005), and the tree search SAT solver SAT4J (which has 4 categorical and 7 numerical parameters; http://www.sat4j.org). They compared the respective algorithm's default performance, the performance of the CALIBRA system (Adenso-Diaz & Laguna, 2006), and the performance of BasicILS and FocusedILS. Out of the four configuration scenarios studied, FocusedILS significantly outperformed CALIBRA on two and performed better on average on the third. For the fourth one (configuring SAT4J), CALIBRA was not applicable due to the categorical parameters, while FocusedILS significantly outperformed BasicILS.

Overall, automated parameter optimization using ParamILS achieved substantial improvements over the previous default settings: GLS$^+$ was sped up by a factor > 360 (tuned parameters found solutions of better quality in 10 seconds than the default found in one hour), SAPS by factors of 8 and 130 on SAPS-QWH and SAPS-SW, respectively, and SAT4J by a factor of 11.

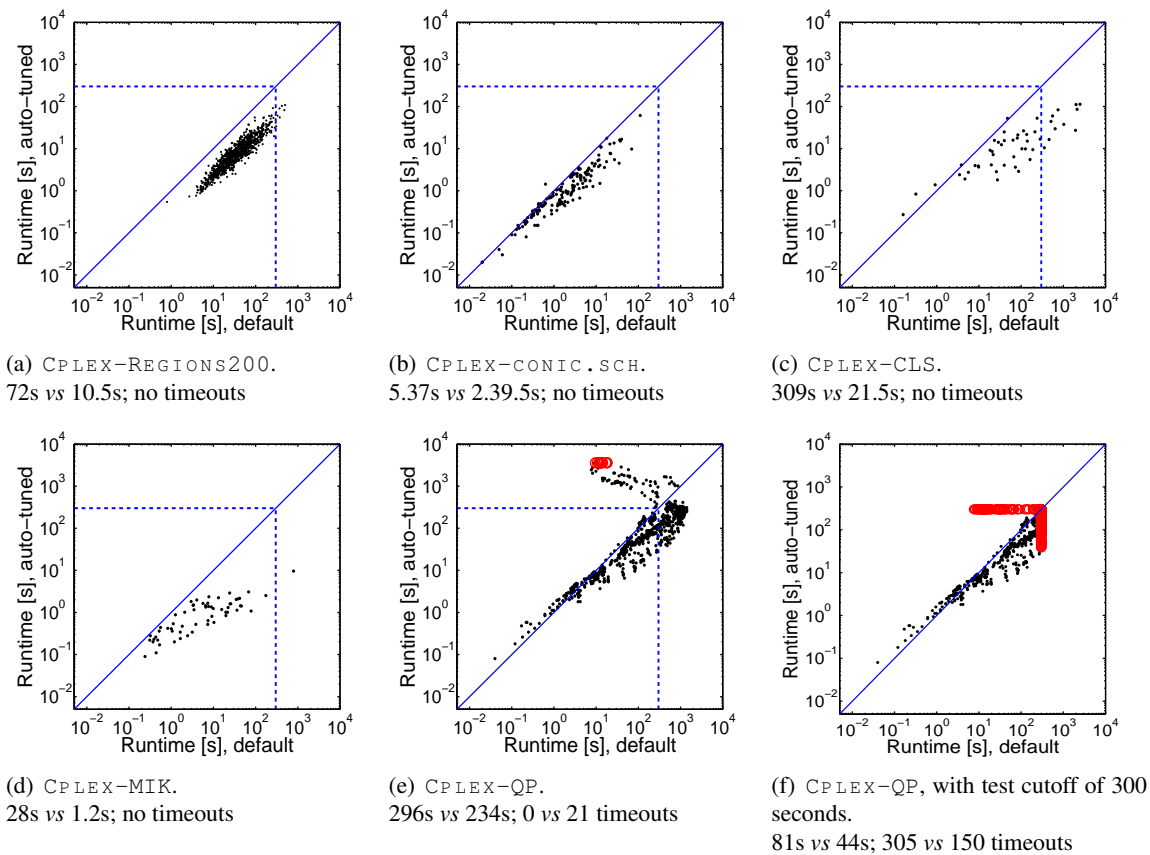

(a) CPLEX-REGIONS200.
72s *vs* 10.5s; no timeouts

(b) CPLEX-CONIC.SCH.
5.37s *vs* 2.39.5s; no timeouts

(c) CPLEX-CLS.
309s *vs* 21.5s; no timeouts

(d) CPLEX-MIK.
28s *vs* 1.2s; no timeouts

(e) CPLEX-QP.
296s *vs* 234s; 0 *vs* 21 timeouts

(f) CPLEX-QP, with test cutoff of 300 seconds.
81s *vs* 44s; 305 *vs* 150 timeouts

Figure 6: Comparison of default *vs* automatically-determined parameter configuration for our five CPLEX configuration scenarios. Each dot represents one test instance; time-outs (after one CPU hour) are denoted by red circles. The blue dashed line at 300 CPU seconds indicates the cutoff time of the target algorithm used during the configuration process. The subfigure captions give mean runtimes for the instances solved by both of the configurations (default *vs* optimized), as well as the number of timeouts for each.

## 8.2 Configuration of Spear for Industrial Verification Problems

Hutter et al. (2007) applied ParamILS to a specific "real-world" application domain: configuring the 26 parameters of the tree-search DPLL solver SPEAR to minimize its mean runtime on a set of practical verification instances. In particular, they considered two sets of industrial problem instances, bounded model-checking (BMC) instances from Zarpas (2005) and software verification (SWV) instances generated by the CALYSTO static checker (Babić & Hu, 2007).

The instances from both problem distributions exhibited a large spread in hardness for SPEAR. For the SWV instances, the default configuration solved many instances in milliseconds but failed to solve others in days. This was despite the fact that SPEAR was specifically developed for this type of instances, that its developer had generated the problem instances himself (and thus had intimate domain knowledge), and that a week of manual performance tuning had been expended in order to optimize the solver's performance.

SPEAR was first configured for good general performance on industrial SAT instances from previous SAT competitions. This already led to substantial improvements over the default perfor-mance in the 2007 SAT competition.[7] While the SPEAR default solved 82 instances and ranked 17th in the first round of the competition, an automatically configured version solved 93 instances and ranked 8th, and a further optimized version solved 99 instances, ranking 5th (above MiniSAT). The speedup factors due to this general optimization were 20 and 1.3 on the SWV and BMC datasets, respectively.

Optimizing on the specific instance sets yielded further, and much larger improvements (a factor of over 500 for SWV and 4.5 for BMC). Most encouragingly, the best parameter configuration found for the software verification instances did not take longer than 20 seconds to solve *any* of the SWV problem instances (compared to multiple timeouts after a CPU day for the original default values).

Key to good performance in that application was to perform multiple independent runs of FocusedILS, and to select the found configuration with best training performance (as also done in Sections 6.1 and 7 of this article).

### 8.3 Configuration of SATenstein

KhudaBukhsh, Xu, Hoos and Leyton-Brown (2009 ) used ParamILS to perform automatic algorithm design in the context of stochastic local search algorithms for SAT. Specifically, they introduced a new framework for local search SAT solvers, called SATenstein, and used ParamILS to choose good instantiations of the framework for given instance distributions. SATenstein spans three broad categories of SLS-based SAT solvers: WalkSAT-based algorithms, dynamic local search algorithms and $G^2WSAT$ variants. All of these are combined in a highly parameterized framework solver with a total of 41 parameters and $4.82 \cdot 10^{12}$ unique instantiations.

FocusedILS was used to configure SATenstein on six different problem distributions, and the resulting solvers were compared to eleven state-of-the-art SLS-based SAT solvers. The results showed that the automatically configured versions of SATenstein outperformed all of the eleven state-of-the-art solvers in all six categories, sometimes by a large margin.

The SATENSTEIN work clearly demonstrated that automated algorithm configuration methods can be used to construct *new* algorithms by combining a wide range of components from existing algorithms in novel ways, and thereby go beyond simple "parameter tuning". Due to the low level of manual work required by this approach, we believe this *automated design of algorithms from components* will become a mainstream technique in the development of algorithms for hard combinatorial problems.

Key to the successful application of FocusedILS for configuring SATENSTEIN was the careful selection of homogeneous instance distributions, most instances of which could be solved within a comparably low cutoff time of 10 seconds per run. Again, the configuration with the best training quality was selected from ten parallel independent runs of FocusedILS per scenario.

### 8.4 Self-Configuration of ParamILS

As a heuristic optimization procedure, ParamILS is itself controlled by a number of parameters: the number of random configurations, $r$, to be sampled at the beginning of search; the perturbation strength, $s$; and the probability of random restarts, $p_{restart}$. Furthermore, our aggressive capping mechanism makes use of an additional parameter: the bound multiplier, $bm$. Throughout this article, we have used the manually-determined default values $\langle r, s, p_{restart}, bm \rangle = \langle 10, 3, 0.01, 2 \rangle$.

---

7. See http://www.cril.univ-artois.fr/SAT07. SPEAR was not allowed to participate in the second round of this competition since its source code is not publicly available.

In recent work (see Section 8.2 of Hutter, 2009), we evaluated whether FocusedILS's performance could be improved by using ParamILS to automatically find a better parameter configuration. In this *self-configuration* task, configuration scenarios play the role of instances, and the configurator to be optimized plays the role of the target algorithm. To avoid confusion, we refer to this configurator as the *target configurator*. Here, we set fairly short configuration times of one CPU hour for the target configurator. However, this was still significantly longer than the cutoff times we used in any of our other experiments, such that parallelization turned out to be crucial to finish the experiment in a reasonable amount of time. Because BasicILS is easier to parallelize than FocusedILS, we chose BasicILS(100) as the meta-configurator.

Although the notion of having an algorithm configurator configure itself was intriguing, in this case, it turned out to only yield small improvements. Average performance improved for four out of the five scenarios and degraded for the remaining one. However, none of the differences was statistically significant.

### 8.5 Other Applications of ParamILS

Thachuk, Shmygelska and Hoos (2007 ) used BasicILS in order to determine performance-optimizing parameter settings of a new replica-exchange Monte Carlo algorithm for protein folding in the 2D-HP and 3D-HP models.[8] Even though their algorithm has only four parameters (two categorical and two continuous), BasicILS achieved substantial performance improvements. While the manually-selected configurations were biased in favour of either short or long protein sequences, BasicILS found a configuration which consistently yielded good mean runtimes for all types of sequences. On average, the speedup factor achieved was approximately 1.5, and for certain classes of protein sequences up to 3. While all manually-selected configurations performed worse than the previous state-of-the-art algorithm for this problem on some instances, the robust parameter configurations selected by BasicILS yielded uniformly better performance.

In very recent work, Fawcett, Hoos and Chiarandini (2009) used several variants of ParamILS (including a version that has been slightly extended beyond the ones presented here) to design a modular stochastic local search algorithm for the post-enrollment course timetabling problem. They followed a design approach that used automated algorithm configuration in order to explore a large design space of modular and highly parameterised stochastic local search algorithms. This quickly led to a solver that placed third in Track 2 of the 2nd International Timetabling Competition (ITC2007) and subsequently produced an improved solver that is shown to achieve consistently better performance than the top-ranked solver from the competition.

## 9. Related Work

Many researchers before us have been dissatisfied with manual algorithm configuration, and various fields have developed their own approaches for automatic parameter tuning. We start this section with the most closely-related work—approaches that employ direct search to find good parameter configurations—and then describe other methods. Finally, we discuss work on related problems, such as finding the best parameter configuration or algorithm on a per-instance basis, and approaches that adapt their parameters during an algorithm's execution (see also Hoos, 2008, for further related work on automated algorithm design).

---

8. BasicILS was used, because FocusedILS had not yet been developed when that study was conducted.

## 9.1 Direct Search Methods for Algorithm Configuration

Approaches for automated algorithm configuration go back to the early 1990s, when a number of systems were developed for *adaptive problem solving*. One of these systems is Composer (Gratch & Dejong, 1992), which performs a hill-climbing search in configuration space, taking moves if enough evidence has been gathered to render a neighbouring configuration statistically significantly better than the current configuration. Composer was successfully applied to improving the five parameters of an algorithm for scheduling communication between a collection of ground-based antennas and spacecrafts (Gratch & Chien, 1996).

Around the same time, the MULTI-TAC system was introduced by Minton (1993, 1996). MULTI-TAC takes as input generic heuristics, a specific problem domain, and a distribution over problem instances. It adapts the generic heuristics to the problem domain and automatically generates domain-specific LISP programs implementing them. A beam search is then used to choose the best LISP program where each program is evaluated by running it on a fixed set of problem instances sampled from the given distribution.

Another search-based approach that uses a fixed training set was introduced by Coy et al. (2001). Their approach works in two stages. First, it finds a good parameter configuration $\theta_i$ for each instance $I_i$ in the training set by a combination of experimental design (full factorial or fractional factorial) and gradient descent. Next, it combines the parameter configurations $\theta_1, \ldots, \theta_N$ thus determined by setting each parameter to the average of the values taken in all of them. Note that this averaging step restricts the applicability of the method to algorithms with only numerical parameters.

A similar approach, also based on a combination of experimental design and gradient descent, using a fixed training set for evaluation, is implemented in the CALIBRA system of Adenso-Diaz and Laguna (2006). CALIBRA starts by evaluating each parameter configuration in a full factorial design with two values per parameter. It then iteratively homes in to good regions of parameter configuration space by employing fractional experimental designs that evaluate nine configurations around the best performing configuration found so far. The grid for the experimental design is refined in each iteration. Once a local optimum is found, the search is restarted (with a coarser grid). Experiments showed CALIBRA's ability to find parameter settings for six target algorithms that matched or outperformed the respective originally-proposed parameter configurations. Its main drawback is the limitation to tuning numerical and ordinal parameters, and to a maximum of five parameters. When we first introduced ParamILS, we performed experiments comparing its performance against CALIBRA (Hutter et al., 2007). These experiments are reviewed in Section 8.1.

Terashima-Marín et al. (1999) introduced a genetic algorithm for configuring a constraint satisfaction algorithm for large-scale university exam scheduling. They constructed and configured an algorithm that works in two stages and has seven configurable categorical parameters. They optimized these choices with a genetic algorithm for each of 12 problem instances, and for each of them found a configuration that improved performance over a modified Brelaz algorithm. However, note that they performed this optimization separately for each instance. Their paper did not quantify how long these optimizations took, but stated that "Issues about the time for delivering solutions with this method are still a matter of research".

Work on automated parameter tuning can also be found in the numerical optimization literature. In particular, Audet and Orban (2006) proposed the mesh adaptive direct search algorithm. Designed for purely continuous parameter configuration spaces, this algorithm is guaranteed to converge to a *local* optimum of the cost function. Parameter configurations were evaluated on a fixed set of large unconstrained regular problems from the CUTEr collection, using as optimization objectives runtime and number of function evaluations required for solving a given problem instance. Performance improvements of around 25% over the classical configuration of four continuous parameters of interior point methods were reported.

Algorithm configuration is a stochastic optimization problem, and there exists a large body of algorithms designed for such problems (see, e.g., Spall, 2003). However, many of the algorithms in the stochastic optimization literature require explicit gradient information and are thus inapplicable to algorithm configuration. Some algorithms approximate the gradient from function evaluations only (e.g., by finite differences), and provably converge to a local minimum of the cost function under mild conditions, such as continuity. Still, these methods are primarily designed to deal with numerical parameters and only find local minima. We are not aware of any applications of general purpose algorithms for stochastic optimization to algorithm configuration.

## 9.2 Other Methods for Algorithm Configuration

Sequential parameter optimization (SPO) (Bartz-Beielstein, 2006) is a model-based parameter optimization approach based on the Design and Analysis of Computer Experiments (DACE; see, e.g., Santner, Williams & Notz, 2003), a prominent approach in statistics for blackbox function optimization. SPO starts by running the target algorithm with parameter configurations from a Latin hypercube design on a number of training instances. It then builds a response surface model based on Gaussian process regression and uses the model's predictions and predictive uncertainties to determine the next parameter configuration to evaluate. The metric underlying the choice of promising parameter configurations is the expected improvement criterion used by Jones, Schonlau and Welch (1998). After each algorithm run, the response surface is refitted, and a new parameter configuration is determined based on the updated model. In contrast to the previously-mentioned methods, SPO does not use a fixed training set. Instead, it starts with a small training set and doubles its size whenever a parameter configuration is determined as incumbent that has already been incumbent in a previous iteration. A recent improved mechanism resulted in a more robust version, SPO$^+$ (Hutter, Hoos, Leyton-Brown & Murphy, 2009). The main drawbacks of SPO and its variants, and in fact of the entire DACE approach, are its limitation to continuous parameters and to optimizing performance for single problem instances, as well as its cubic runtime scaling in the number of data points.

Another approach is based on adaptations of racing algorithms in machine learning (Maron & Moore, 1994) to the algorithm configuration problem. Birattari et al. (2002; 2004) developed a procedure dubbed F-Race and used it to configure various stochastic local search algorithms. F-Race takes as input an algorithm $\mathcal{A}$, a finite set of algorithm configurations $\Theta$, and an instance distribution $\mathcal{D}$. It iteratively runs the target algorithm with all "surviving" parameter configurations on a number of instances sampled from $\mathcal{D}$ (in the simplest case, each iteration runs all surviving configurations on one instance). A configuration is eliminated from the race as soon as enough statistical evidence is gathered against it. After each iteration, a non-parametric Friedman test is used to check whether there are significant differences among the configurations. If this is the case, the inferior configurations are eliminated using a series of pairwise tests. This process is iterated until only one configuration survives or a given cutoff time is reached. Various applications of F-Race have demonstrated very good performance (for an overview, see Birattari, 2004). However, since at the start of the procedure all candidate configurations are evaluated, this approach is limited to situations in which the space of candidate configurations can practically be enumerated. In fact, published ex-periments with F-Race have been limited to applications with only around 1200 configurations. A recent extension presented by Balaprakash et al. (2007) iteratively performs F-Race on subsets of parameter configurations. This approach scales better to large configuration spaces, but the version described by Balaprakash et al. (2007) handles only algorithms with numerical parameters.

### 9.3 Related Algorithm Configuration Problems

Up to this point, we have focused on the problem of finding the best algorithm configuration for an entire set (or distribution) of problem instances. Related approaches attempt to find the best configuration or algorithm on a per-instance basis, or to adapt algorithm parameters during the execution of an algorithm. Approaches for setting parameters on a per-instance basis have been described by Patterson and Kautz (2001), Cavazos and O'Boyle (2006), and Hutter et al. (2006). Furthermore, approaches that attempt to select the best *algorithm* on a per-instance basis have been studied by Leyton-Brown, Nudelman and Shoham (2002), Carchrae and Beck (2005), Gebruers, Hnich, Bridge and Freuder (2005), Gagliolo and Schmidhuber (2006), and Xu, Hutter, Hoos and Leyton-Brown (2008). In other related work, decisions about when to restart an algorithm are made online, during the run of an algorithm (Horvitz, Ruan, Gomes, Kautz, Selman & Chickering, 2001; Kautz, Horvitz, Ruan, Gomes & Selman, 2002; Gagliolo & Schmidhuber, 2007). So-called reactive search methods perform online parameter modifications (Battiti, Brunato & Mascia, 2008). This last strategy can be seen as complementary to our work: even reactive search methods tend to have parameters that remain fixed during the search and can hence be configured using offline approaches such as ParamILS.

### 9.4 Relation to Other Local Search Methods

Since ParamILS performs an iterated local search with a one-exchange neighbourhood, it is very similar in spirit to local search methods for other problems, such as SAT (Selman, Levesque & Mitchell, 1992; Hoos & Stützle, 2005), CSP (Minton, Johnston, Philips & Laird, 1992), and MPE (Kask & Dechter, 1999; Hutter et al., 2005). Since ParamILS is a local search method, existing theoretical frameworks (see, e.g., Hoos, 2002; Mengshoel, 2008), could in principle be used for its analysis. The main factor distinguishing our problem from the ones faced by "standard" local search algorithms is the stochastic nature of our optimization problem (for a discussion of local search for stochastic optimization, see, e.g., Spall, 2003). Furthermore, there exists no compact representation of the objective function that could be used to guide the search. To illustrate this, consider local search for SAT, where the candidate variables to be flipped can be limited to those occurring in currently-unsatisfied clauses. In general algorithm configuration, on the other hand, such a mechanism cannot be used, because the only information available about the target algorithm is its performance in the runs executed so far. While, obviously, other (stochastic) local search methods could be used as the basis for algorithm configuration procedures, we chose iterated local search, mainly because of its conceptual simplicity and flexibility.

## 10. Discussion, Conclusions and Future work

In this work, we studied the problem of automatically configuring the parameters of complex, heuristic algorithms in order to optimize performance on a given set of benchmark instances. We extended our earlier algorithm configuration procedure, ParamILS, with a new capping mechanism and obtained excellent results when applying the resulting enhanced version of ParamILS to two high-performance SAT algorithms as well as to CPLEX and a wide range of benchmark sets.

Compared to the carefully-chosen default configurations of these target algorithms, the parameter configurations found by ParamILS almost always performed much better when evaluated on sets of previously unseen test instances, for some configuration scenarios by as much as two orders of magnitude. The improvements over CPLEX's default parameter configuration are particularly noteworthy, though we do *not* claim to have found a new parameter configuration for CPLEX that is uniformly better than its default. Rather, given a somewhat homogeneous instance set, we find a configuration specific to that set that typically outperforms the default, sometimes by a factor as high as 20. Note that we achieved these results even though we are *not* intimately familiar with CPLEX and its parameters; we chose the parameters to optimize as well as the values to consider based on a single person-day of studying the CPLEX user manual. The success of automated algorithm configuration even under these extreme conditions demonstrates the potential of the approach.

The ParamILS source code and executable are freely available at

<div align="center">

`http://www.cs.ubc.ca/labs/beta/Projects/ParamILS/`,

</div>

along with a quickstart guide and data for the configuration scenarios studied in this article.[9]

In order to apply ParamILS, or other such automated algorithm configuration methods, a practitioner must supply the following ingredients.

- **A parameterized algorithm** $A$ It must be possible to set $A$'s configurable parameters externally, e.g., in a command line call. Often, a search for hard-coded parameters hidden in the algorithm's source code can lead to a large number of additional parameters to be exposed.

- **Domains for the parameters** Algorithm configurators must be provided with the allowable values for each parameter. Depending on the configurator, it may be possible to include additional knowledge about dependencies between parameters, such as the conditional parameters supported by ParamILS. For the use of ParamILS, numerical parameters must be discretized to a finite number of choices. Depending on the type of parameter, a uniform spacing of values or some other spacing, such as uniform on a log scale, is typically reasonable.

- **A set of problem instances** The more homogeneous the problem set of interest is, the better we can expect any algorithm configuration procedure to perform on it. While it is possible to configure an algorithm for good performance on rather heterogeneous instance sets (e.g., on industrial SAT instances, as we did with SPEAR as reported in Section 8.2), the results for homogeneous subsets of interest will improve when we configure on instances from that subset. Whenever possible, the set of instances should be split into disjoint training and test sets in order to safeguard against over-tuning. When configuring on a small and/or heterogeneous benchmark set, ParamILS (or any other configuration procedure) might not find configurations that perform well on an independent test set.

- **An objective function** While we used median performance in our first study on ParamILS (Hutter et al., 2007), we have since found cases where optimizing median performance led to parameter configurations with good median but poor overall performance. In these cases, optimizing for mean performance yielded more robust parameter configurations. However, when optimizing mean performance one has to define the cost for unsuccessful runs. In this article, we have penalized such runs by counting them as ten times the cutoff time. How to deal with unsuccessful runs in a more principled manner is an open research question.

---

9. ParamILS continues to be actively developed; it is currently maintained by Chris Fawcett.

- **A cutoff time for unsuccessful runs** The smaller the cutoff time for each run of the target algorithm is chosen, the more quickly any configuration procedure will be able to explore the configuration space. However, choosing too small a cutoff risks the failure mode we experienced with our `CPLEX-QP` scenario. Recall that there, choosing 300 seconds as a timeout yielded a parameter configuration that was very good when judged with that cutoff time (see Figure 6(f)), but performed poorly for longer cutoffs (see Figure 6(e)). In all of our other experiments, parameter configurations performing well with low cutoff times turned out to scale well to harder problem instances as well. In many configuration scenarios, in fact, we noticed that our automatically-found parameter configurations showed much *better* scaling behaviour than the default configuration. We attribute this to our use of mean runtime as a configuration objective. The mean is often dominated by the hardest instances in a distribution. However, in manual tuning, algorithm developers typically pay more attention to easier instances, simply because repeated profiling on hard instances takes too long. In contrast, a "patient" automatic configurator can achieve better results because it avoids this bias.

- **Computational resources** The amount of (computational) time required for the application of automated algorithm configuration clearly depends on the target application. If the target algorithm takes seconds to solve instances from a homogeneous benchmark set of interest, in our experience a single five-hour configuration run will suffice to yield good results and for some domains we have achieved good results with configuration times as short as half an hour. In contrast, if runs of the target algorithm are slow and only performance with a large cutoff time can be expected to yield good results on the instances of interest, then the time requirements of automated algorithm configuration grow. We also regularly perform multiple parallel configuration runs and pick the one with best *training* performance in order to deal with variance across configuration runs.

Overall, we firmly believe that automated algorithm configuration methods such as ParamILS will play an increasingly prominent role in the development of high-performance algorithms and their applications. The study of such methods is a rich and fruitful research area with many interesting questions remaining to be explored.

In ongoing work, we are currently developing methods that adaptively adjust the domains of integer-valued and continuous parameters during the configuration process. Similarly, we plan to enhance ParamILS with dedicated methods for dealing with continuous parameters that do not require discretization by the user. Another direction for further development concerns the strategic selection of problem instances used during evaluation of configurations and of instance-specific cutoff times used in this context. By heuristically preventing the configuration procedure from spending inordinate amounts of time trying to evaluate poor parameter settings on very hard problem instances, it should be possible to improve its scalability.

We believe that there is significant room for combining aspects of the methods studied here with concepts from related work on this and similar algorithm configuration problems. In particular, we believe it would be fruitful to integrate statistical testing methods—as used, e.g., in F-Race—into ParamILS. Furthermore, we see much potential in the use of response surface models and active learning, and believe these can be combined with our approach. Finally, while the algorithm configuration problem studied in this article is of significant practical importance, there is also much to be gained from studying methods for related problems, in particular, instance-specific algorithm configuration and the online adjustment of parameters during the run of an algorithm.

## Acknowledgments

We thank Kevin Murphy for many helpful discussions regarding this work. We also thank Domagoj Babić, the author of SPEAR, and Dave Tompkins, the author of the UBCSAT SAPS implementation we used in our experiments. We thank the researchers who provided the instances or instance generators used in our work, in particular Gent et al. (1999), Gomes and Selman (1997), Leyton-Brown et al. (2000), Babić and Hu (2007), Zarpas (2005), Le Berre and Simon (2004), Aktürk et al. (2007), Atamtürk and Muñoz (2004), Atamtürk (2003), and Andronescu et al. (2007). Lin Xu created the specific sets of `QCP` and `SWGCP` instances we used. Thanks also to Chris Fawcett and Ashique KhudaBukhsh for their comments on a draft of this article. Finally, we thank the anonymous reviewers as well as Rina Dechter and Adele Howe for their valuable feedback. Thomas Stützle acknowledges support from the F.R.S.-FNRS, of which he is a Research Associate. Holger Hoos acknowledges support through NSERC Discovery Grant 238788.